\documentclass[10pt,twocolumn,letterpaper]{article}

\usepackage{cvpr}
\usepackage{times}
\usepackage{epsfig}
\usepackage{graphicx}
\usepackage{amsmath}
\usepackage{amssymb}
\usepackage{gensymb}

\usepackage[pagebackref=true,breaklinks=true,letterpaper=true,colorlinks,bookmarks=false]{hyperref}

\cvprfinalcopy 

\def\cvprPaperID{****} 

\ifcvprfinal\fi
\begin{document}

\title{Accurate Trajectory Prediction for Autonomous Vehicles}
\author{Michael Diodato
\qquad
Yu Li
\qquad
Antonia Lovjer
\qquad
Minsu Yeom
\qquad
Albert Song
\qquad
Yiyang Zeng\\
Abhay Khosla
\qquad
Benedikt Schifferer
\qquad
Manik Goyal
\\
Iddo Drori\\
\\
Columbia University\\
School of Engineering and Applied Science\\
}



\maketitle

\begin{abstract}
Predicting vehicle trajectories, angle and speed is important for safe and comfortable driving. We demonstrate the best predicted angle, speed, and best performance overall winning the top three places of the ICCV 2019 Learning to Drive challenge. Our key contributions are (i) a general neural network system architecture which embeds and fuses together multiple inputs by encoding, and decodes multiple outputs using neural networks, (ii) using pre-trained neural networks for augmenting the given input data with segmentation maps and semantic information, and (iii) leveraging the form and distribution of the expected output in the model. We make our models and code publicly available \cite{sourcecode}.
\end{abstract}


\section{Introduction}
\label{sec:introduction}

\begin{figure}
\begin{center}
    \includegraphics[width=1\linewidth]{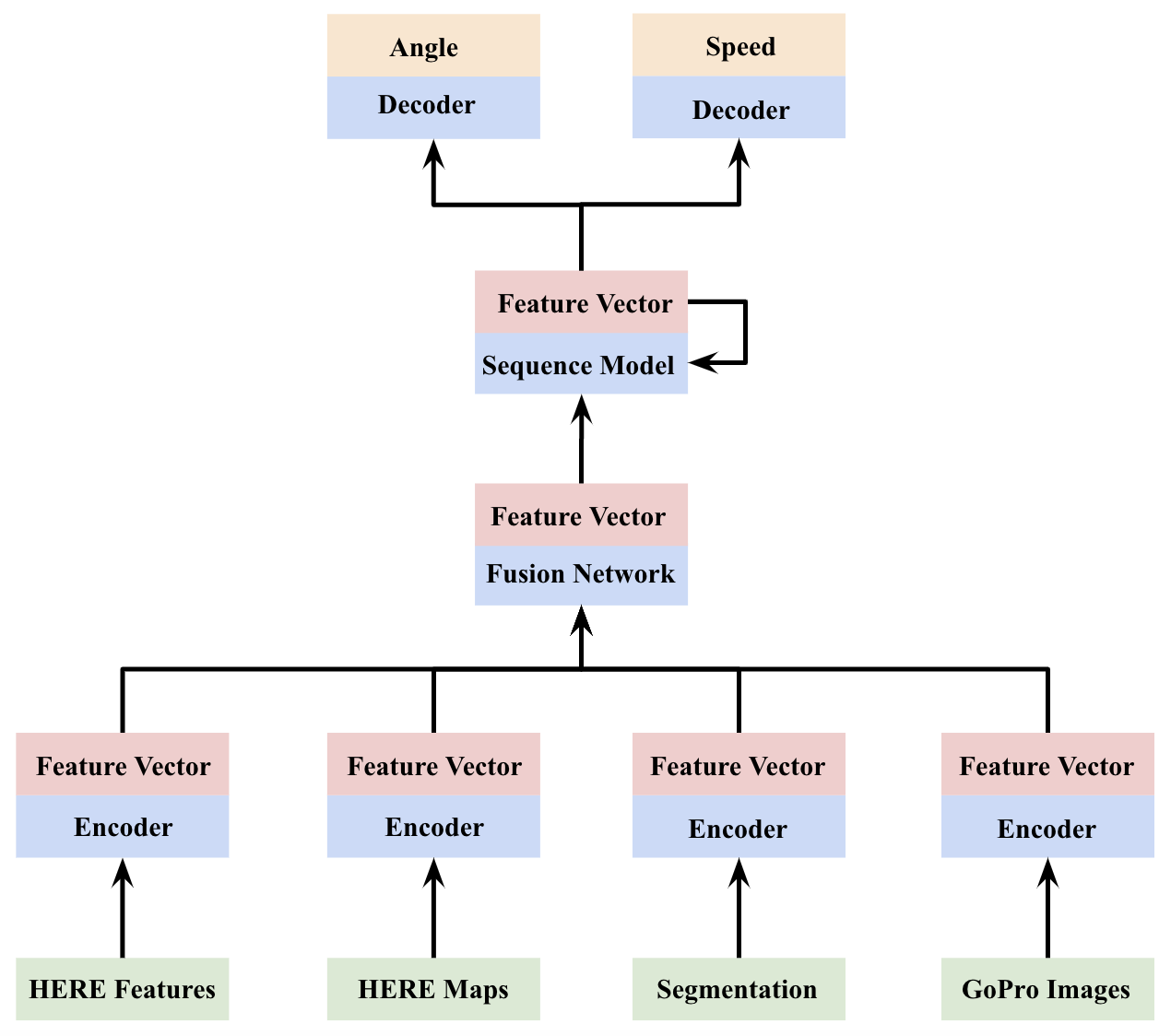}
    \caption{Network architecture: inputs in green, neural networks in blue, intermediate outputs in red, and final outputs in orange. A cycle denotes a sequence model with multiple time steps.}
    \label{fig:nnarchitecture}
\end{center}
\end{figure}

Self driving cars have moved from driving in the desert terrain, spearheaded by the DARPA Grand Challenge, through highways, and into populated cities. Traditionally, a mediated perception approach was used where the entire scene is parsed to make a decision by solving sub-problems \cite{DeepDriving}. Deep learning architectures consist of end-to-end models \cite{bojarski2016end, xu2017end, fernando2017going, bansal2018chauffeurnet} that directly map input images to driving actions \cite{end2end}, predicting trajectories by supervised learning. Our three key contributions are:
\begin{enumerate}
    \item End-to-end neural network system architectures that (i) encode multiple input representations (camera images, segmentation maps, semantic maps, semantic features), followed by (ii) a neural network that fuses these different modalities together, and finally (iii) separate decoding networks for generating both outputs. A general prototype is shown in Figure \ref{fig:nnarchitecture} and we ensemble several variants of this architecture.
    \item We use pre-trained models for generating segmentation maps of the input images and semantic information. The pre-trained models augment the existing inputs with a rich dataset.
    \item We leverage the fact that both speed and steering angles are smooth functions to improve the predictions.
\end{enumerate}

\begin{figure*}
    \begin{center}
        \includegraphics[width=0.195\linewidth]{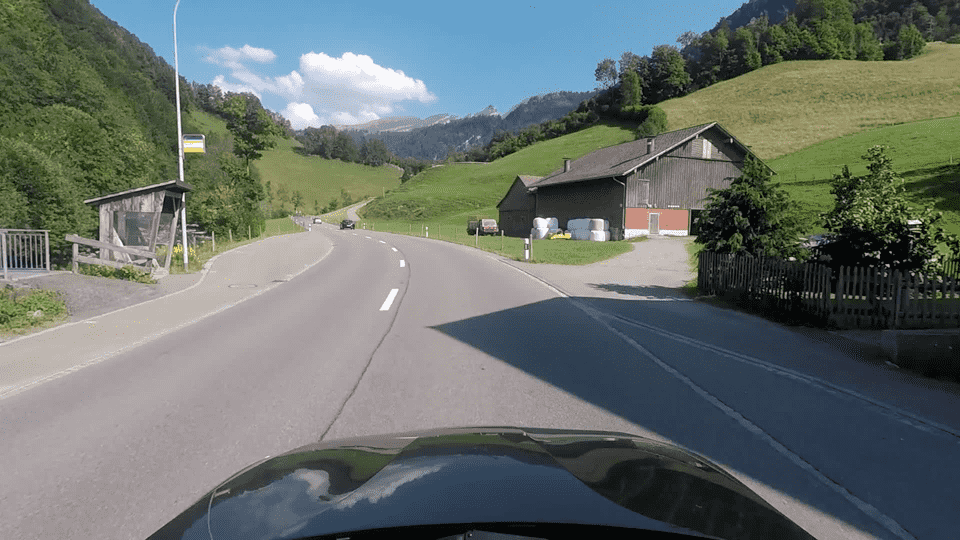}
        \includegraphics[width=0.195\linewidth]{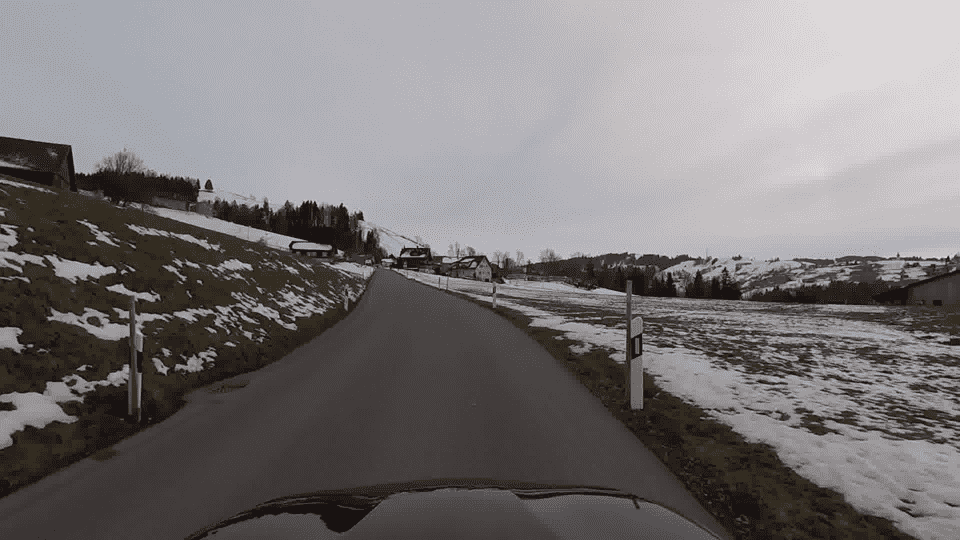}
        \includegraphics[width=0.195\linewidth]{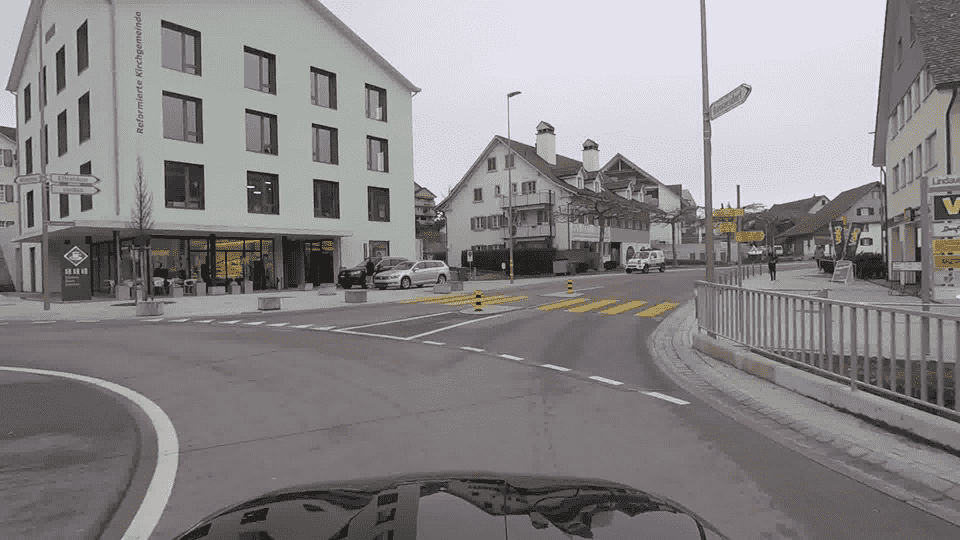}
        \includegraphics[width=0.195\linewidth]{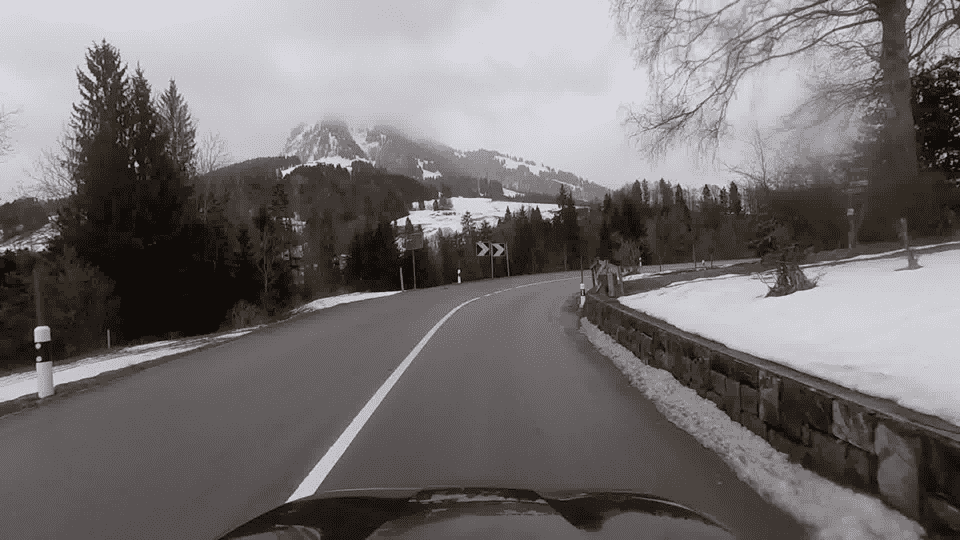}
        \includegraphics[width=0.195\linewidth]{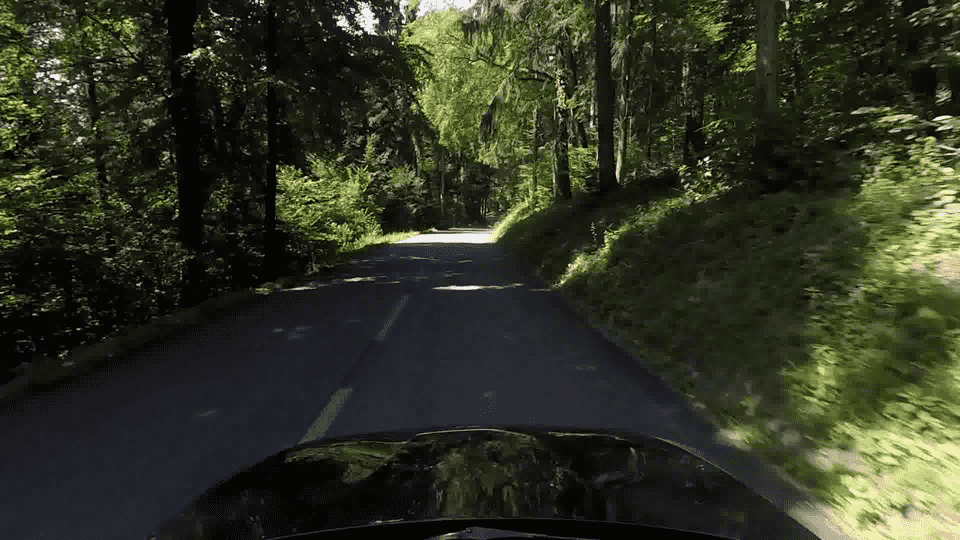}
        \includegraphics[width=0.195\linewidth]{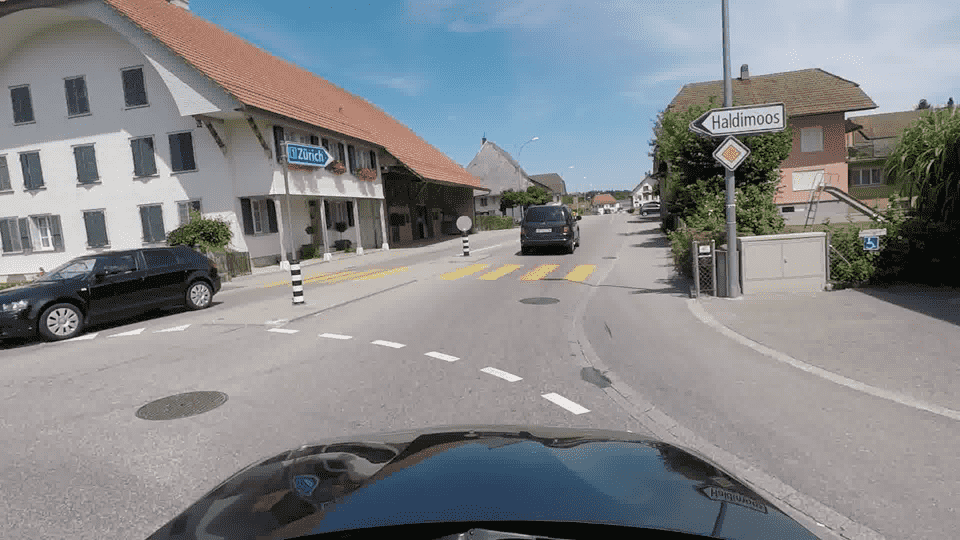}
        \includegraphics[width=0.195\linewidth]{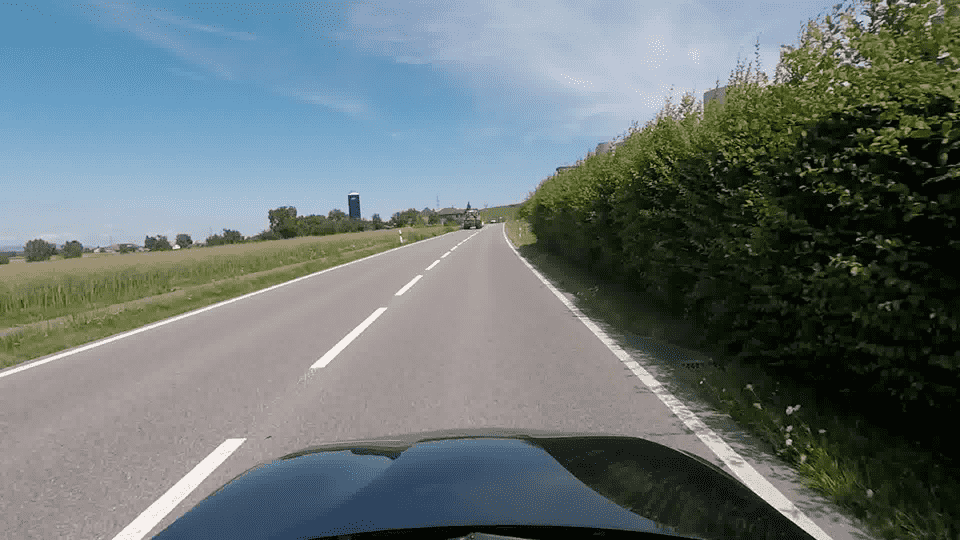}
        \includegraphics[width=0.195\linewidth]{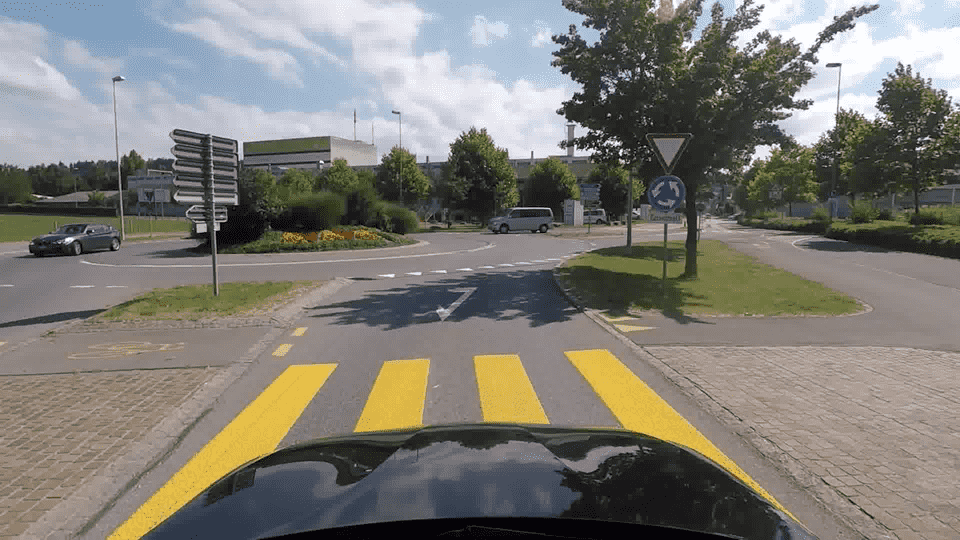}
        \includegraphics[width=0.195\linewidth]{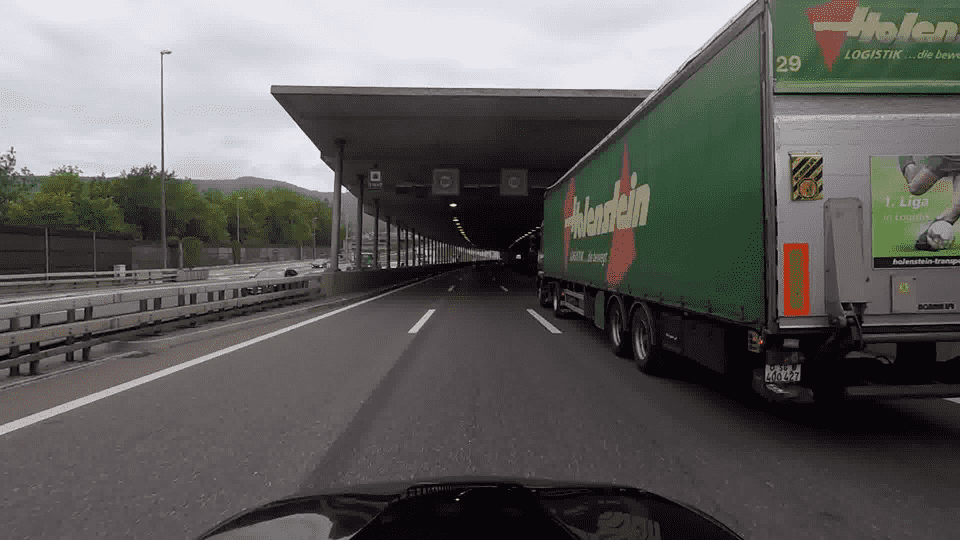}
        \includegraphics[width=0.195\linewidth]{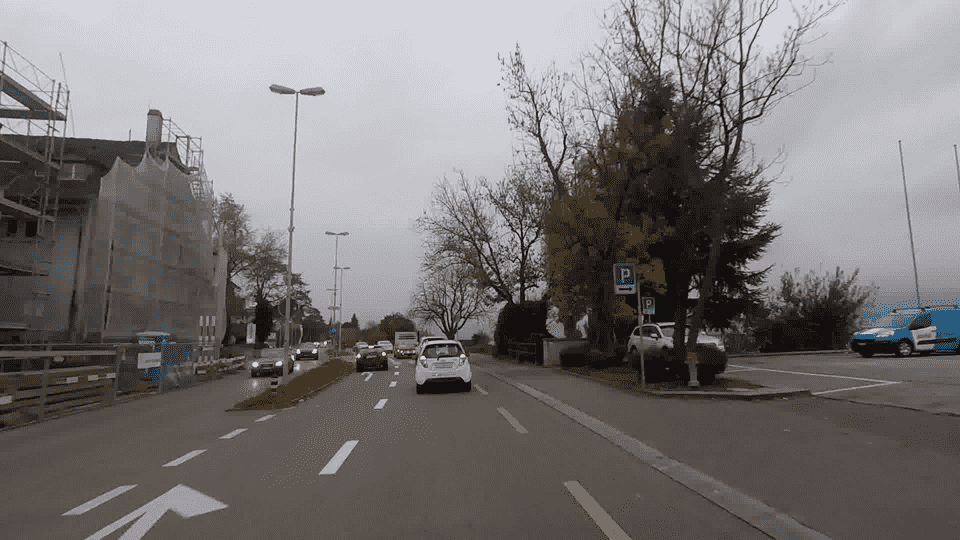}
        \end{center}
    \caption{Samples of front facing training images. The training data set consists of about 1.6 million images from each of the four sides of a car driving through Switzerland and consists of a mixture of cities, highways, and rural areas. The videos were taken by a GoPro Hero 5, and are sampled at a rate of 10 frames per second. Each frame has a resolution of 1920x1080, and the videos are split into chapters of 5 minutes each. The training dataset has 548 chapters from 27 unique routes. In addition to the images, the training dataset also includes visual maps from HERE Technologies at each time, a semantic map that is derived using a Hidden-Markov-Model path matcher, and the steering wheel angle and vehicle speed.}
    \label{fig:train}
\end{figure*}

\begin{figure*}
    \begin{center}
        \includegraphics[width=0.195\linewidth]{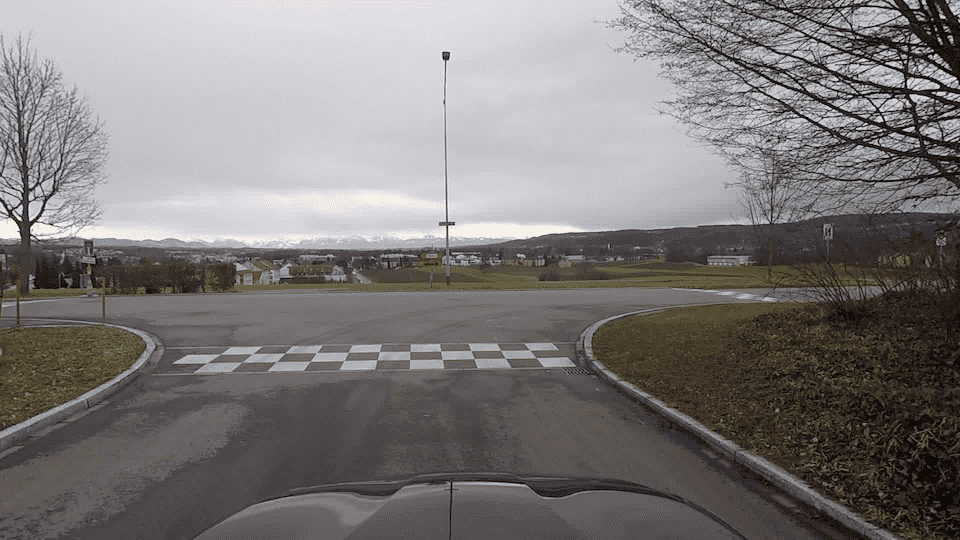}
        \includegraphics[width=0.195\linewidth]{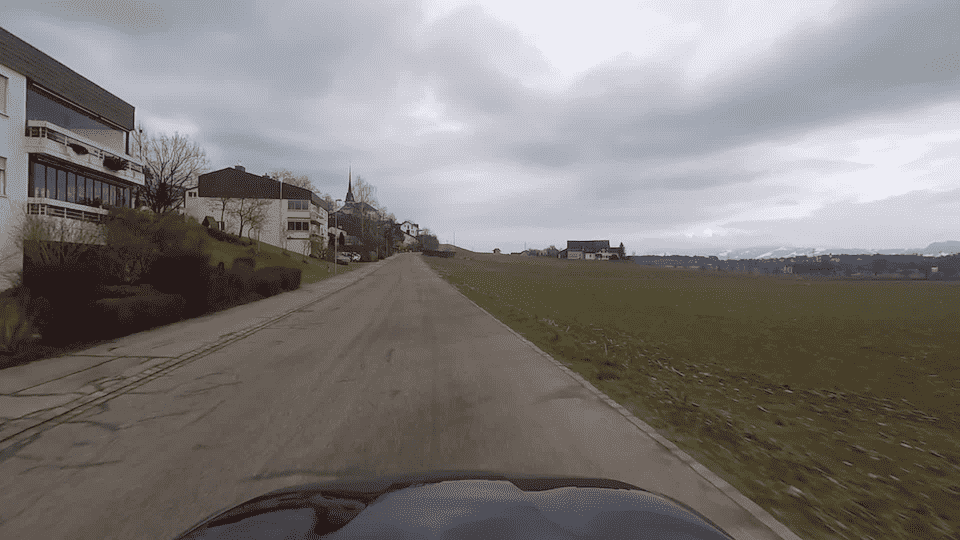}
        \includegraphics[width=0.195\linewidth]{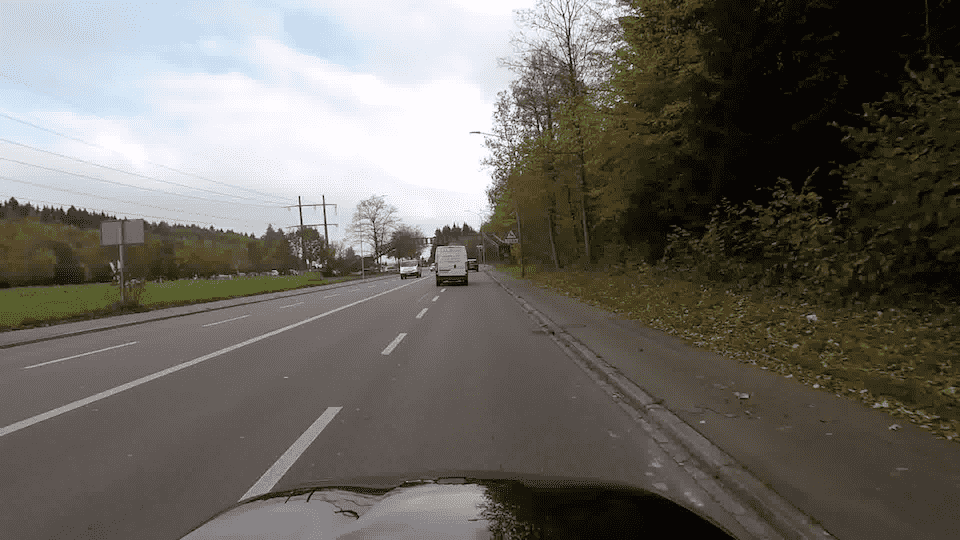}
        \includegraphics[width=0.195\linewidth]{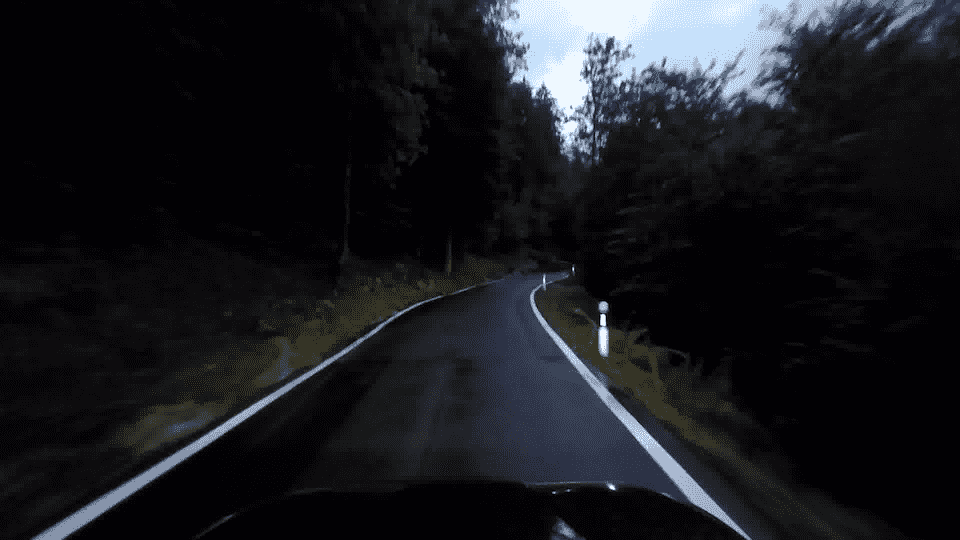}
        \includegraphics[width=0.195\linewidth]{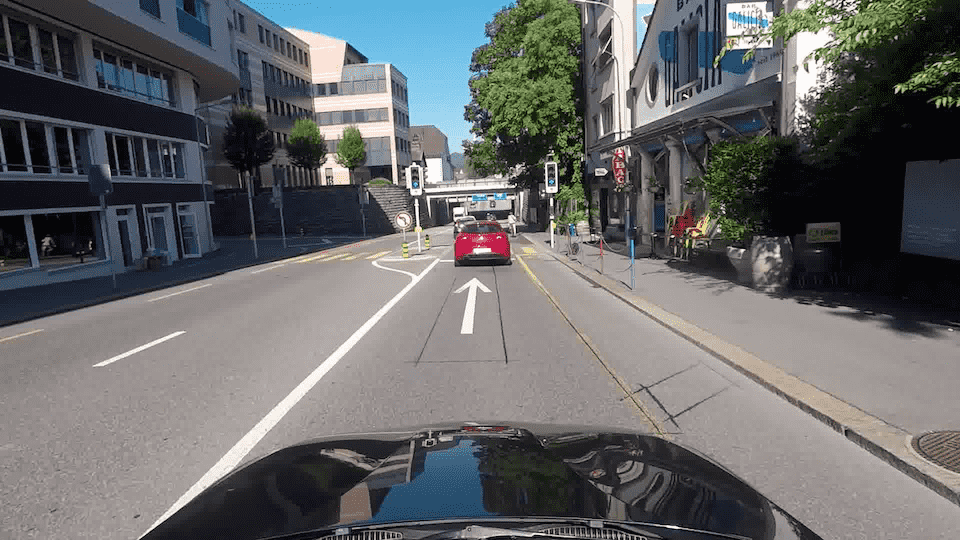}
        \includegraphics[width=0.195\linewidth]{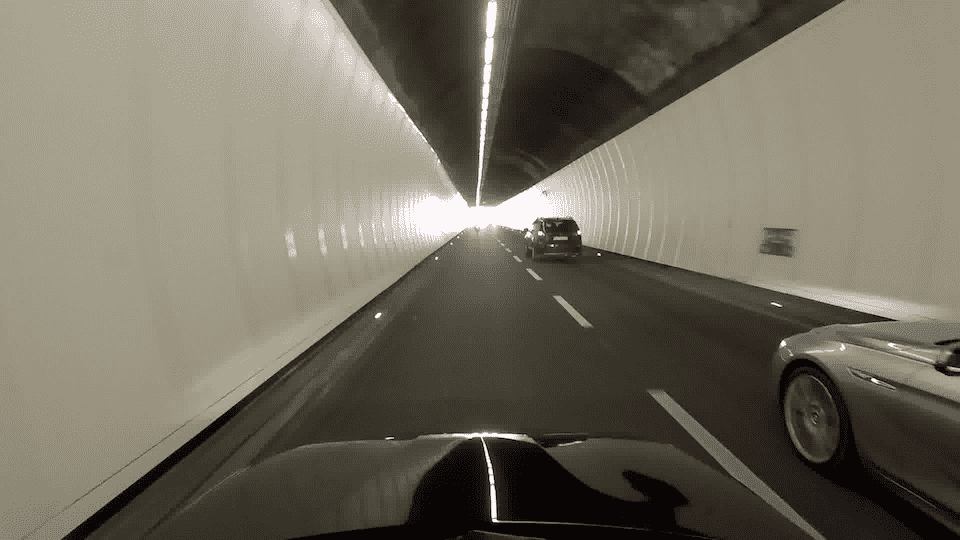}
        \includegraphics[width=0.195\linewidth]{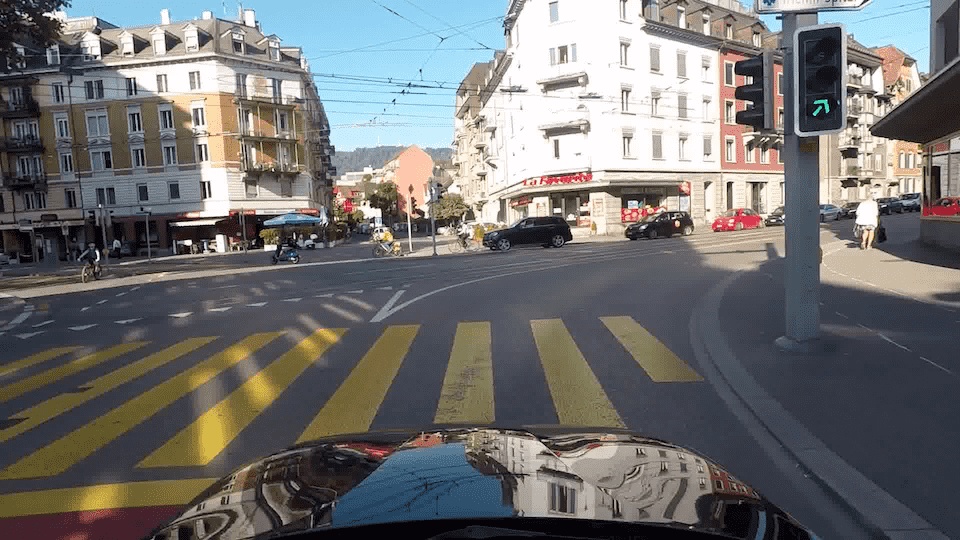}
        \includegraphics[width=0.195\linewidth]{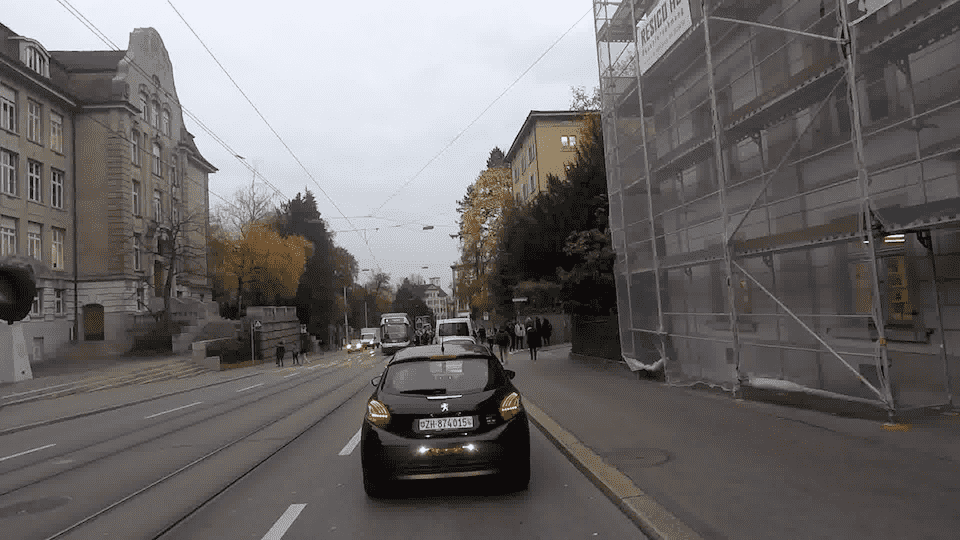}
        \includegraphics[width=0.195\linewidth]{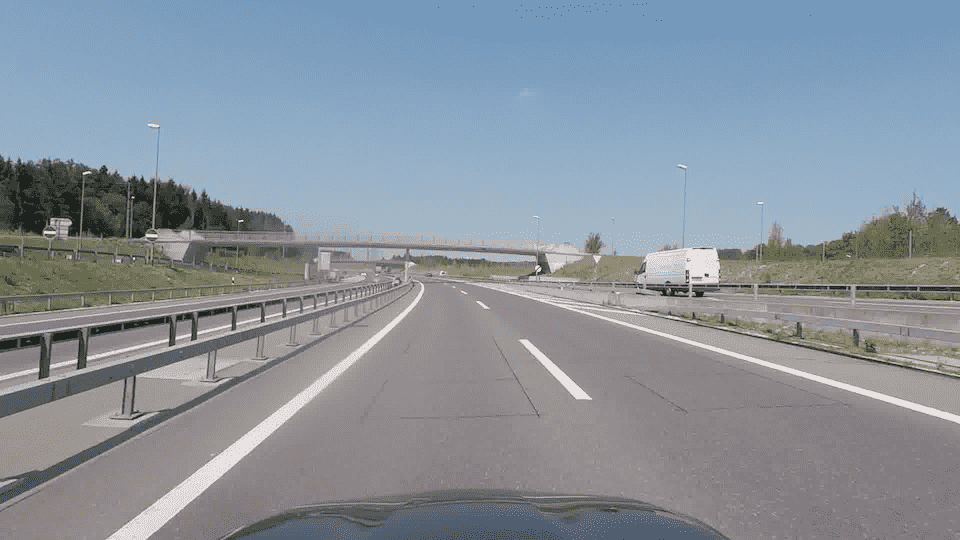}
        \includegraphics[width=0.195\linewidth]{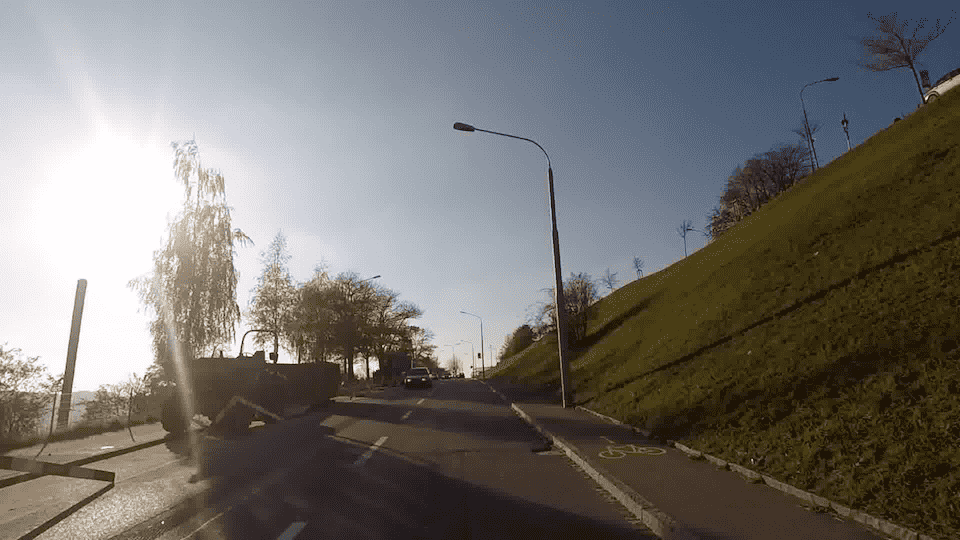}
        \end{center}
    \caption{Sample of front facing test images. The test data set consists of 289,663 images from each of the four sides of a vehicle driving through Switzerland and consists of a mixture of cities, highways, and rural areas. The test image dataset consists of 98 chapters from 27 unique routes.}
    \label{fig:test}
\end{figure*}

We evaluate our results on the Drive360 dataset \cite{hecker2018end} in the Learning to Drive Challenge. We win the top three positions overall, compared with state-of-the-art methods ranking below 10th position \cite{hecker2019learning}. The dataset consists of videos of around 55 hours of recorded driving in Switzerland, along with their associated driving speed and steering angle. Sample images from the training set are shown in Figure \ref{fig:train} and sample test images are shown in Figure \ref{fig:test}. The dataset consists of about 2 million images recorded using a GoPro Hero5 facing the front, left, right, and rear of the vehicle. The dataset also includes images of visual maps from HERE technologies and their corresponding semantic map features. Specifically, the semantic map consists of 21 fields as well as GPS latitude, longitude, and precision. All images and data are sampled at 10 frames per-second. The data is separated into 5 minute chapters. In total, there are 682 chapters for 27 different routes. The data is randomly sampled into 548 chapters for training, 36 chapters for validation, and 98 chapters for testing.


We show that jointly predicting steering angle and vehicle speed is improved by using segmentation maps and data augmentation. Motivated by the use of semantic segmentation models for self driving vehicles \cite{xu2017end, hou2019learning}, we concatenate the segmentation maps with the images as the input instead of using the maps as an additional learning objective. We augment the data set by applying transformations including mirroring, adjusting brightness, and geometric transformations. We ensemble three neural network architectures to achieve the best performance. 

\section{Related Work}
End-to-end deep learning models have been trained to predict steering angle given only the front camera view \cite{bojarski2016end}. As humans have a wider perceptional field than the front camera, 360-degree view datasets have been collected with additional route planner maps \cite{hecker2018end, hecker2019learning}. Neural network models have been trained end-to-end using these 360-degree views which are specifically useful for navigating cities and crossing intersections. Map data has been demonstrated to improve the steering angle prediction accuracy \cite{hecker2018end}. LSTM models achieve good results predicting steering angle by taking into account long range dependencies \cite{fernando2017going}. Another improvement is the usage of event cameras instead of traditional cameras, which capture moving edges \cite{maqueda2018event}. Since predicting steering angle alone is insufficient for self-driving cars, a multi-task learning framework is used to predict both speed and steering angle in an end-to-end fashion \cite{yang2018end}. Adding a segmentation loss to fully connected layers has been shown to improve overall performance by learning a better feature representation \cite{xu2017end}. Recently, ChaeffeurNet \cite{bansal2018chauffeurnet} predicted trajectories using a mid-level controller allowing to predict once and transfer the results to many vehicles types, avoiding the need to retrain a model for every different vehicle type.

%

\begin{figure}
    \begin{center}
        \includegraphics[width=0.495\linewidth]{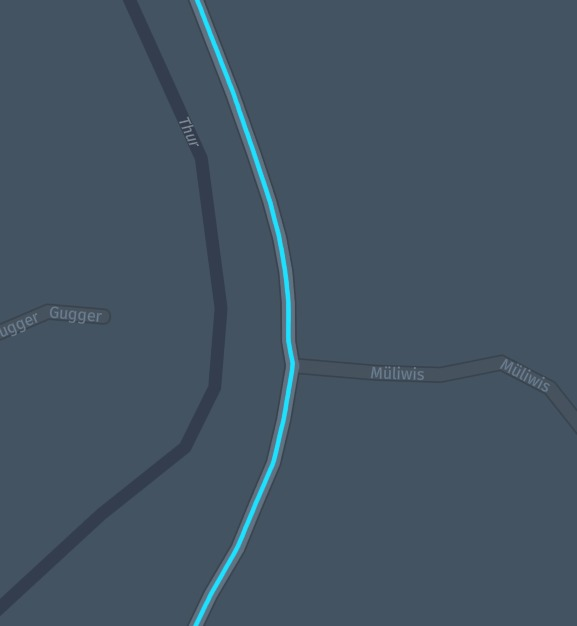}
        \includegraphics[width=0.495\linewidth]{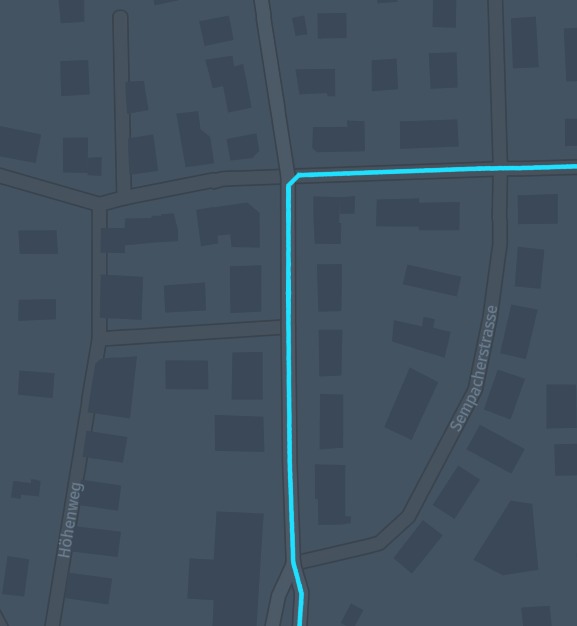}
        \end{center}
    \caption{Sample visual maps from HERE Technologies for each set of GoPro images in the training and test sets at a 10 frames per second sampling rate.}
    \label{fig:here}
\end{figure}

\begin{figure}
    \begin{center}
        \includegraphics[width=0.477\linewidth]{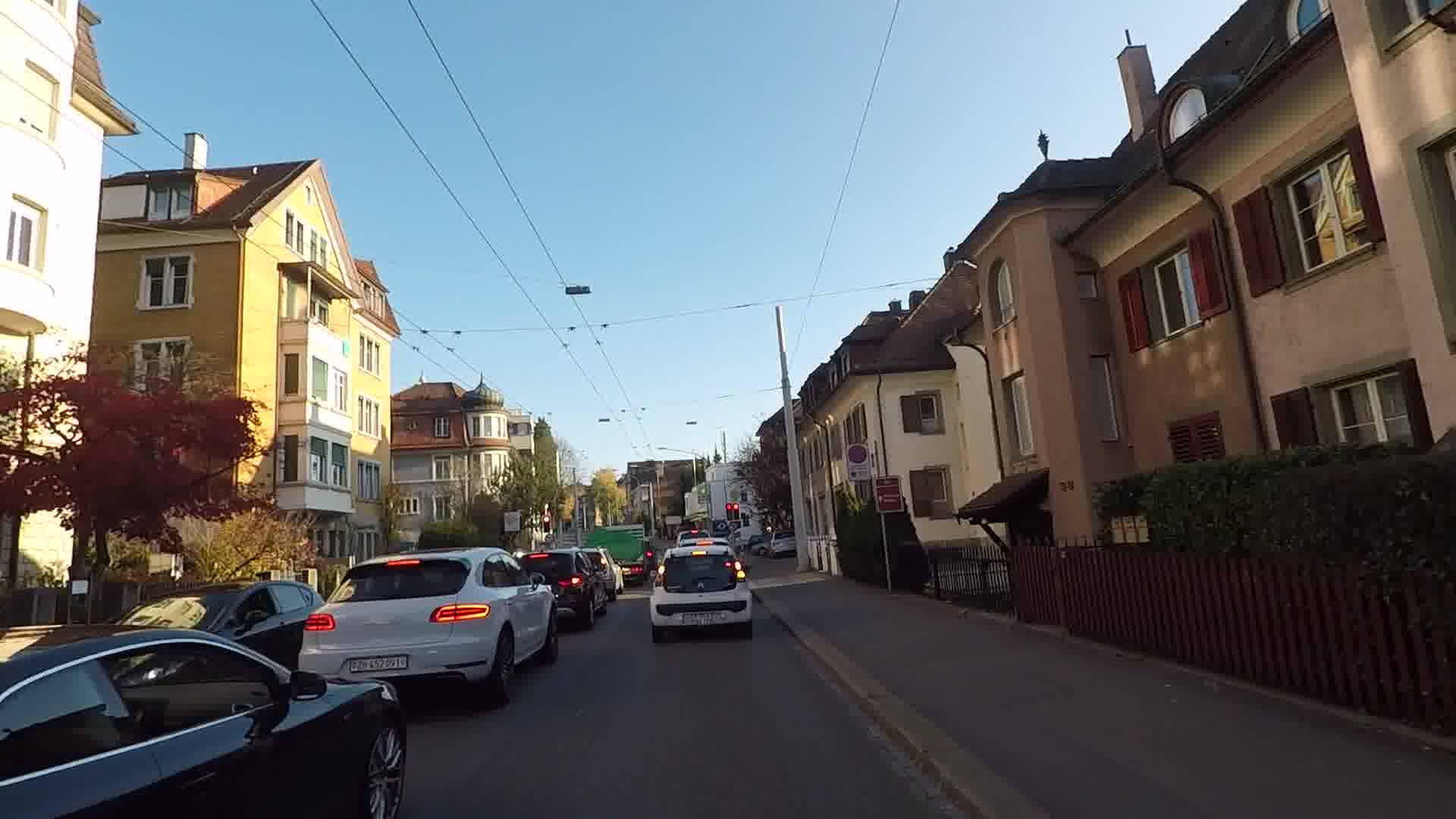}
        \includegraphics[width=0.477\linewidth]{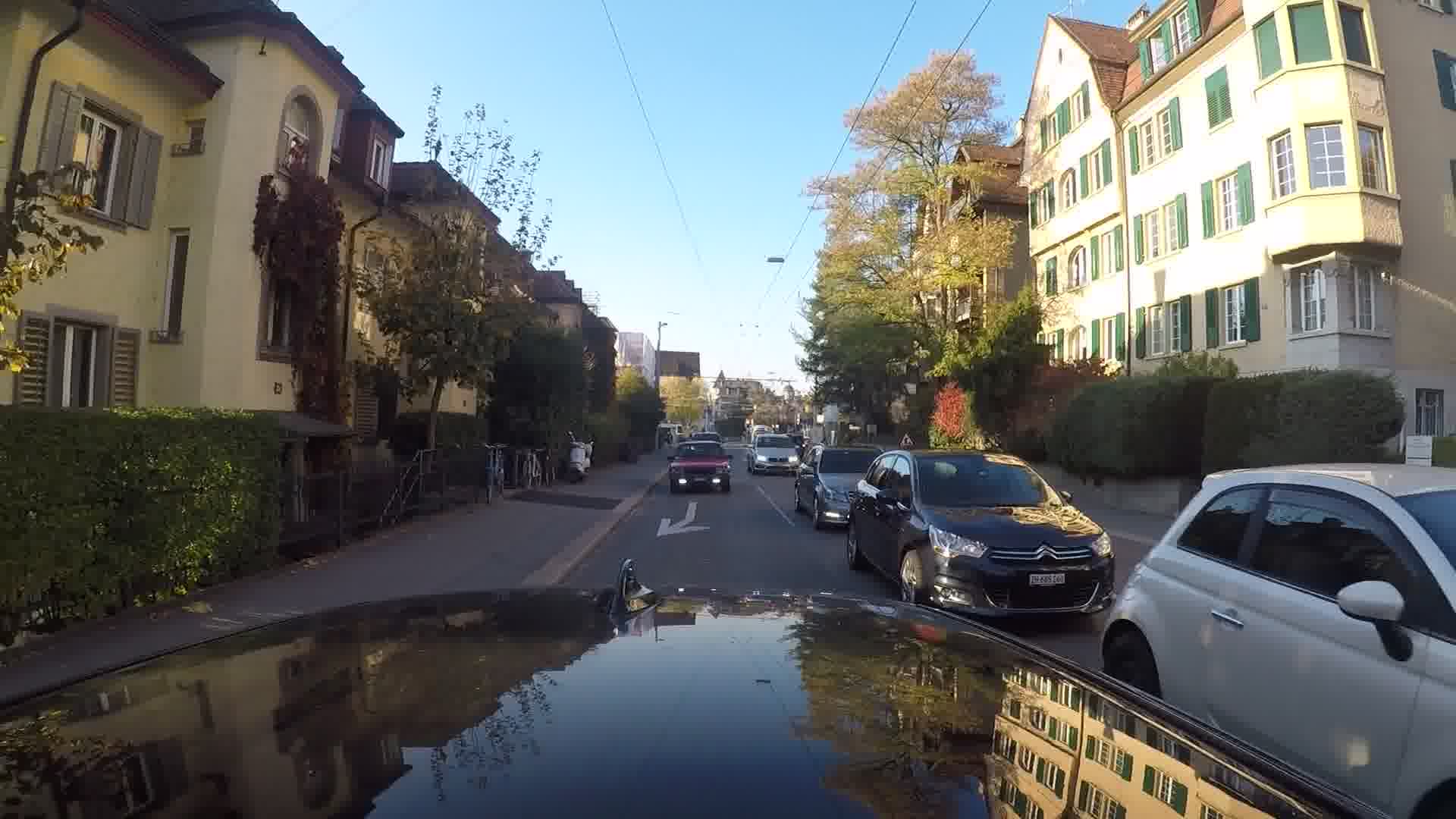}
        \includegraphics[width=0.477\linewidth]{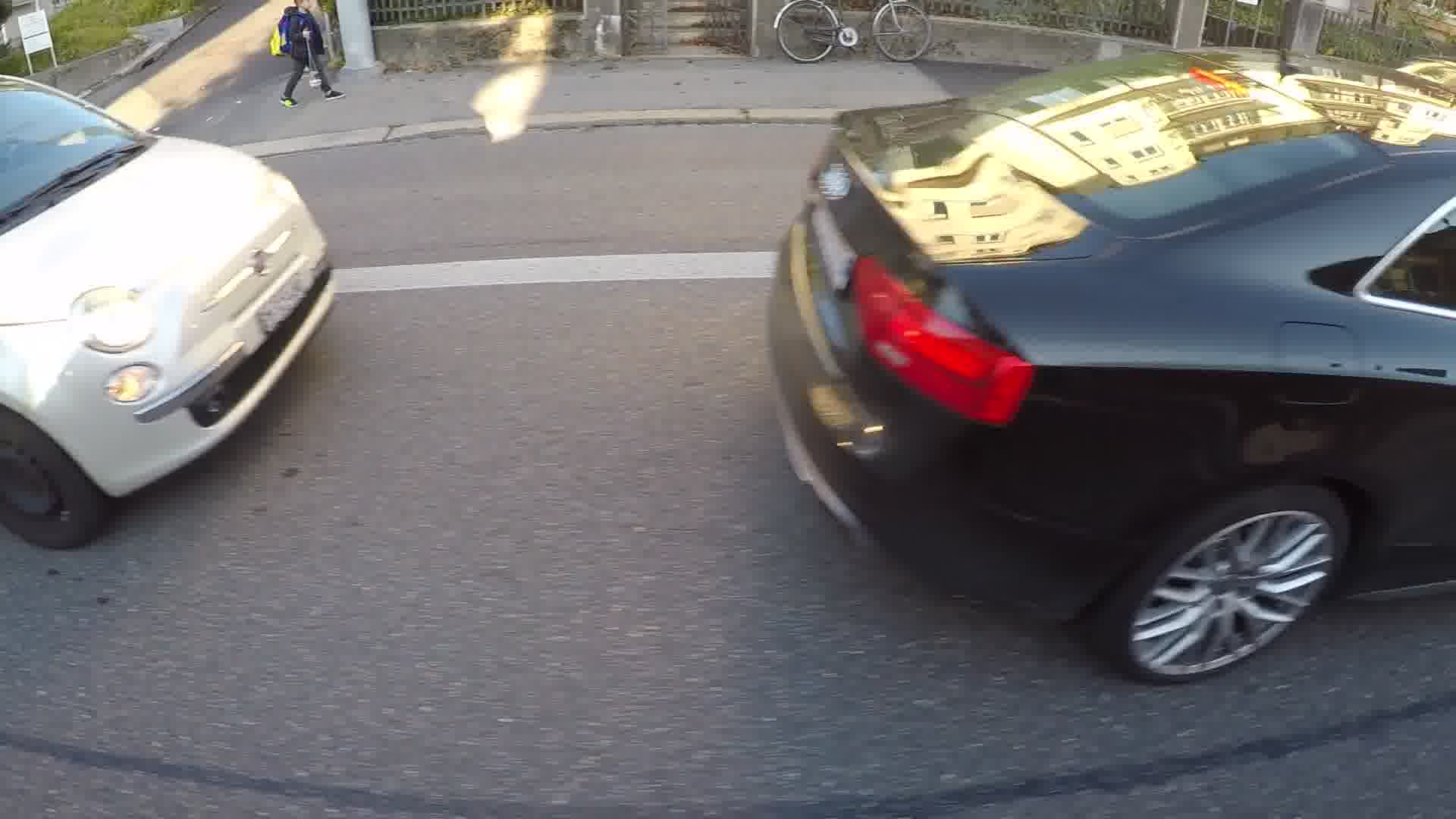}
        \includegraphics[width=0.477\linewidth]{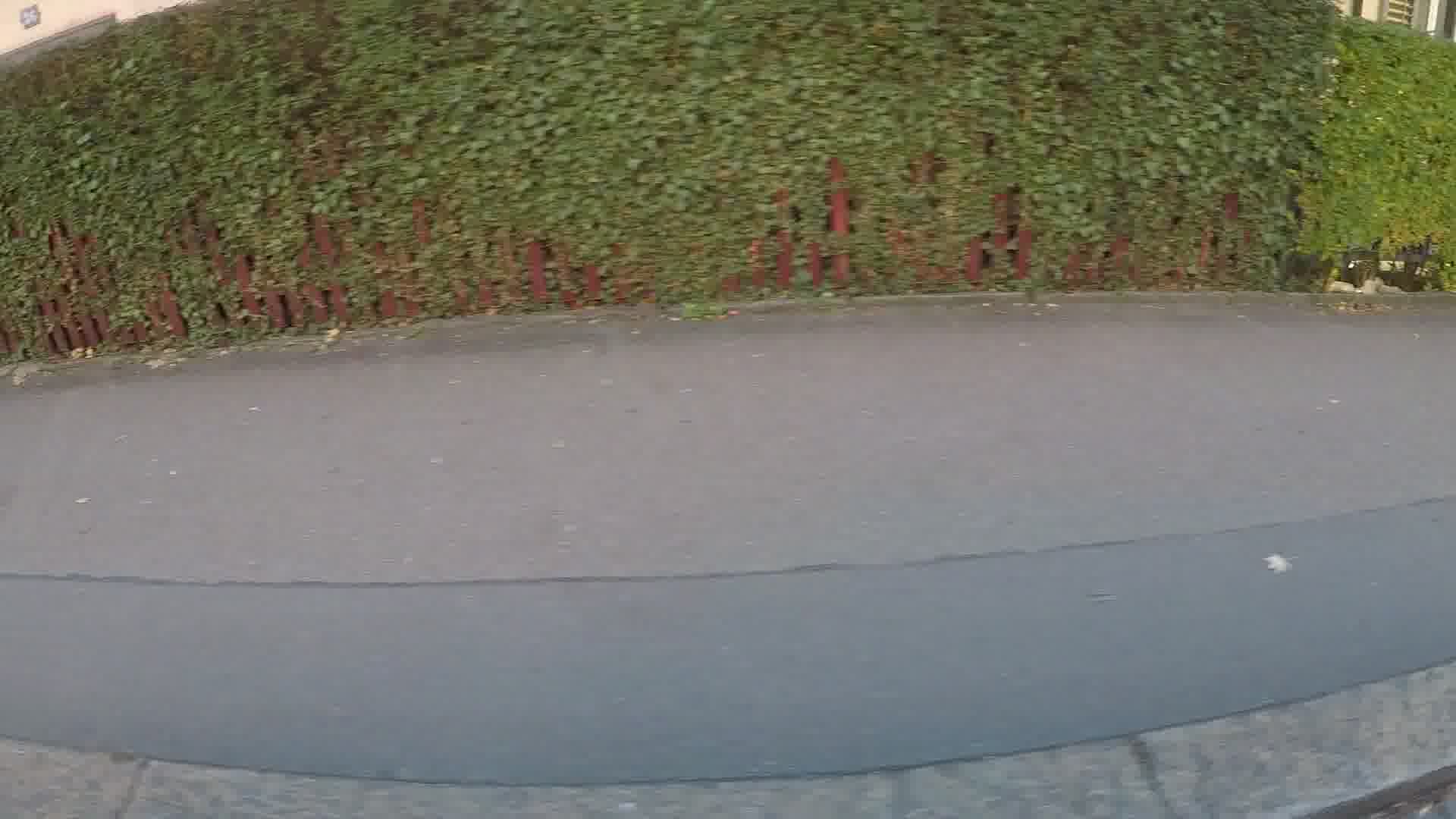}
        \end{center}
    \caption{Sample front, rear, left, and right images for the same time and location. The corresponding HERE map for the instance is shown in Figure \ref{fig:here} (right). Each sampling of the dataset includes images in the four directions and a HERE map. }
    \label{fig:all}
\end{figure}
\section{Methods}
\label{sec:methods}

We pre-process the data by conducting down-sampling, normalization, semantic map imputation and data augmentation to allow for fast experimentation and improved results. We also augment the dataset using segmentation masks derived from a pre-trained model. The different modalities and pre-processed data are fed into two types of models, one of which uses the images and semantic map information and pre-trained ResNets, and another of which takes as input the images as well as segmentation masks, using a combination of pre-trained and fully-trained networks. 

\paragraph{Down-sampling.}
For efficient experimentation we down-sample the dataset in both space and time as shown in Table \ref{tab:sampling}. Although this initial pre-processing is time consuming, down-sampling enables us to train models faster by orders of magnitude, reducing computation time from days to minutes. The initial image size is also prohibitive due to memory limitations on our GPU's and significantly decreases the speed at which we train the model. In both cases an ensemble method is used to combine results and provide the highest achieving results in the competition.

\begin{table}
\small
\begin{center}
\begin{tabular}{l|llll}
Dataset & Full & Sample 1 & Sample 2 & Sample 3 \\
\hline
Res.  & 1920x1080 & 640x360  & 320x180  & 160x90   \\
Train & 1,600,000 & 160,000     & 80,000      & 40,000\\
Val.  & 106,000   & 10,600    & 5,300     & 2,650    \\
Test  & 290,000      & 29,000      & 14,500    & 7,250 \\
\hline
\end{tabular}
\end{center}
\caption{Data sampling in both space and time for efficiency.}
\label{tab:sampling}
\end{table}

\paragraph{Normalization.} The images are normalized in all models according to the  mean and standard deviation. Similarly, the steering angle and speed are normalized for training using means and standard deviation.

\paragraph{Semantic map.} We use 20 of the numerical features including latitude, longitude, speed limit from a navigation map, free flow speed (average driving speed based on underlying road geometry), headings and road indices at intersections, as well as road distance to next signal, yield, pedestrian crossing, and intersection.
We imputed missing values by zeros, avoiding using the mean value across the chapter or other placeholder, to avoid using future data. For one model, we add in 27 additional dummy variables for each folder an image was located. Although the semantic map has information on location, the information from the folder gave another view that we consider possibly helpful to the training process. 


\paragraph{Data augmentation.}
We augmented the dataset by applying several transformations to the training data (i) random horizontal flips of frontal images with probability 0.5 (steering angle is multiplied by -1 in order to offset this flip, (ii) random changes to image brightness by a factor between 0.2 and 0.75 with probability 0.1, and (iii) random translations and rotations of frontal image with probability 0.25.

\paragraph{Segmentation mask.}
We used a pre-trained segmentation model \cite{semantic_cvpr19, Cordts_2016_CVPR} to generate segmentation masks for each of the images in the dataset. The original pre-trained model contains 34 classes of which we consider only 19 which include relevant objects (such as road, car, parking, wall, etc.) that influence the steering angle \cite{Cordts_2016_CVPR}. 

\subsection{Architecture}
We experimented with three different network architectures which won 1st, 2nd, and 3rd places in the competition.

\subsubsection{Model 1 network architecture (1st place)}

\begin{figure}
\begin{center}
    \includegraphics[width=0.6\linewidth]{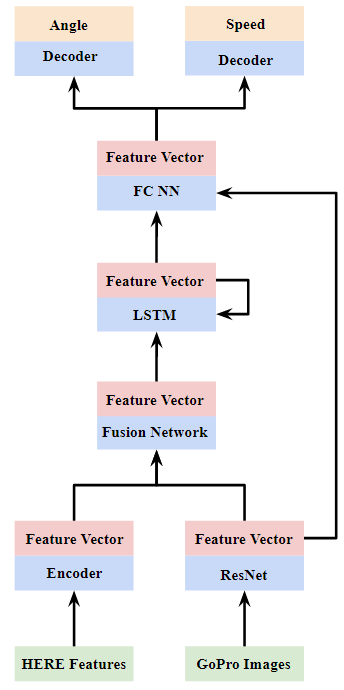}
    \caption{Model 1: The network consists of a pre-trained ResNet and fully connected network that feeds into an LSTM model. This and the output of the ResNet model on the current image are fed into an angle and speed regressor. Each regressor consists of 3 blocks of a linear layer, a ReLU activation, and a 10\% droput layer. The hidden layers have dimensions 1024, 512, and 256.} 
    \label{fig:architecture}
\end{center}
\end{figure}

Figure \ref{fig:architecture} shows Model 1's network architecture. The inputs consist of the front facing GoPro \cite{gopro} images and the semantic map from HERE Technologies \cite{here}. We include the current and previous frame in each iteration, which are 0.4 seconds apart. The images are fed into a pre-trained ResNet34 model or a ResNet152 model. We feed the semantic map into a fully connected network with two hidden layers of dimensions 256 and 128 with ReLU activation layers. The output from the fully connected and ResNet models are concatenated and fed into a long short-term memory (LSTM) network. The output from the LSTM network and the current information from the semantic map, if used, are concatenated. The LSTM output and the current information is then used as input to both an angle regressor and a speed regressor, which predict the current steering angle and speed.

\subsubsection{Model 2 network architectures (2nd place)}
\begin{figure}
\centering
    \includegraphics[width=0.5\linewidth]{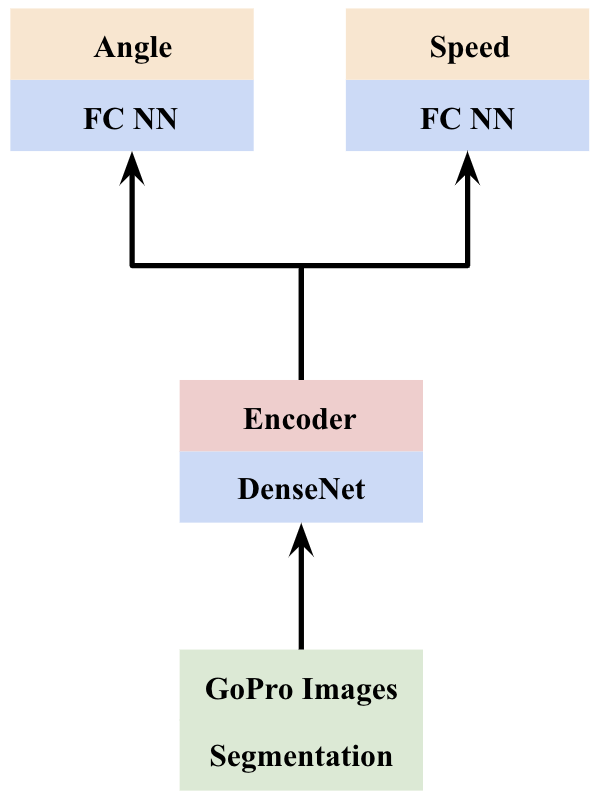}
    \caption{Model 2.1: model 2-single and model 2-stacked share the same architecture, where the only difference is the input. For model 2-single, the input is a single image concatenated with its one-hot encoded segmentation mask. The input dimensions are width $\times$ height $\times$ 23 (3 RGB channels + 20 classes in the segmentation mask). For model 2-stacked, the input is a full sequence of 10 images concatenated together with their segmentation masks. All images and masks are stacked to a width $\times$ height $\times$ 230 (3 RGB channels  + 20 classes)*10. These inputs are fed into a pre-trained DenseNet121. The final layers are two separate feedforward towers - one for speed and for steering angle. Each of the feed forward layers, except of the output layer, are followed by batch normalization and ReLU activation.}
    \label{fig:arch_B}
\end{figure}

\begin{figure}
\centering
    \includegraphics[width=0.6\linewidth]{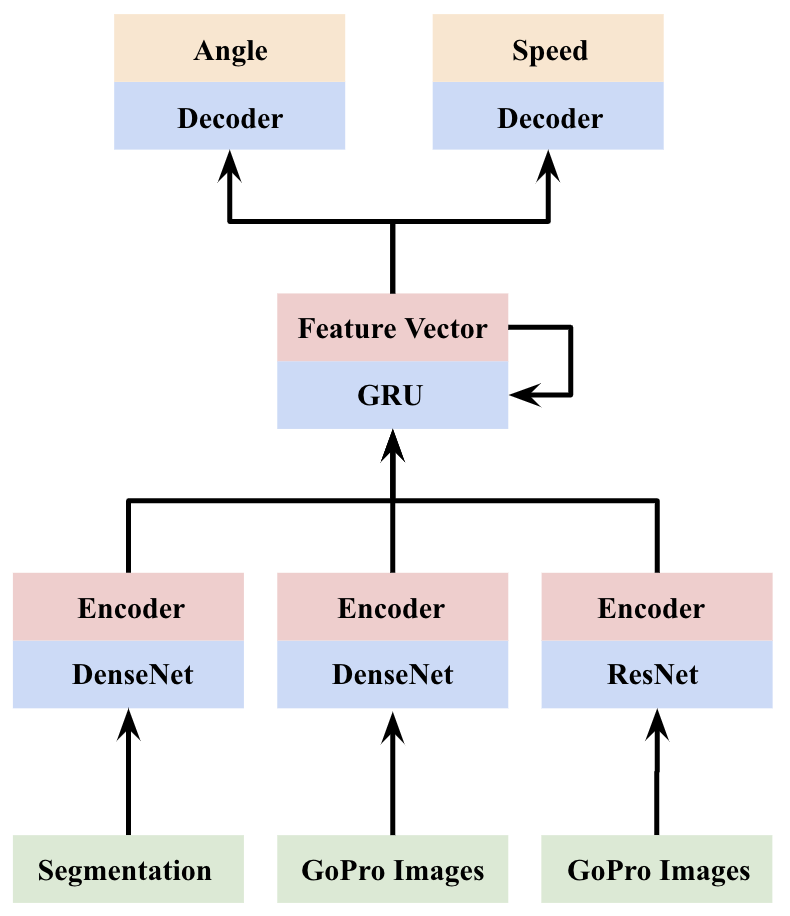}
    \caption{Model 2-sequence: The input is a sequence of 10 images and the corresponding segmentation masks. The images are passed through a pre-trained ResNet34 and a pre-trained DenseNet 121, while the segmentation masks are concatenated with their corresponding image and passed through model 2-signal. For each image and segmentation mask, the hidden features after the convolutional layers are extracted and passed through 2 fully-connected layers. The result is a sequence of features, which is fed into a 3-layer bi-directional GRU with 64 hidden features. The output of the GRU is concatenated with the hidden features of the latest input image and its segmentation mask, which is finally fed into two feed forward towers for speed and angle prediction.}
    \label{fig:arch_C}
\end{figure}

Figures \ref{fig:arch_B} and \ref{fig:arch_C} show two variants of model 2 that differ in their inputs: one taking a single image as input and the other taking a sequence of images as input.

Model 2-single takes as input a single image and its corresponding segmentation mask. The input is passed through a DenseNet121 architecture, followed by decoders for predicting speed and angle implemented using fully connected networks with three dense layers of size 200, 50 and 10 each. In between each dense layer we apply batch normalization a ReLU non-linearity. The final output is a real-valued number which is the predicted speed or steering angle, which is normalized using the mean and standard deviation from the training set. Model 2-stacked takes as input a sequence of 10 images, where each image is concatenated with its corresponding segmentation mask.

Model 2-sequence shown in Figure \ref{fig:arch_C} takes as input a sequence of 10 images, and their corresponding segmentation masks, similar to Model 2-stacked. Each image in the input sequence is passed individually through a pre-trained ResNet34 model and a pre-trained DenseNet201. Additionally, the input images are concatenated with their corresponding segmentation masks and passed through model 2-single. The resulting outputs are concatenated and passed through an intermediate layer which contains two dense layers of dimensions 512 and 128 with dropout. The output is then passed into a bi-directional GRU. This is performed for each input image/mask pair in the sequence, and the output of the GRU is concatenated with the input to the previous fully connected layer which is also passed through a dense block. Finally, this representation is passed into two decoders for speed and steering angle prediction, each consisting of three fully connected layers of dimensions 256, 128 and 32. We ensemble the various model 2 variants.

\subsubsection{Model 3 network architecture (3rd place)}
\begin{figure}
\centering
    \includegraphics[width=0.6\linewidth]{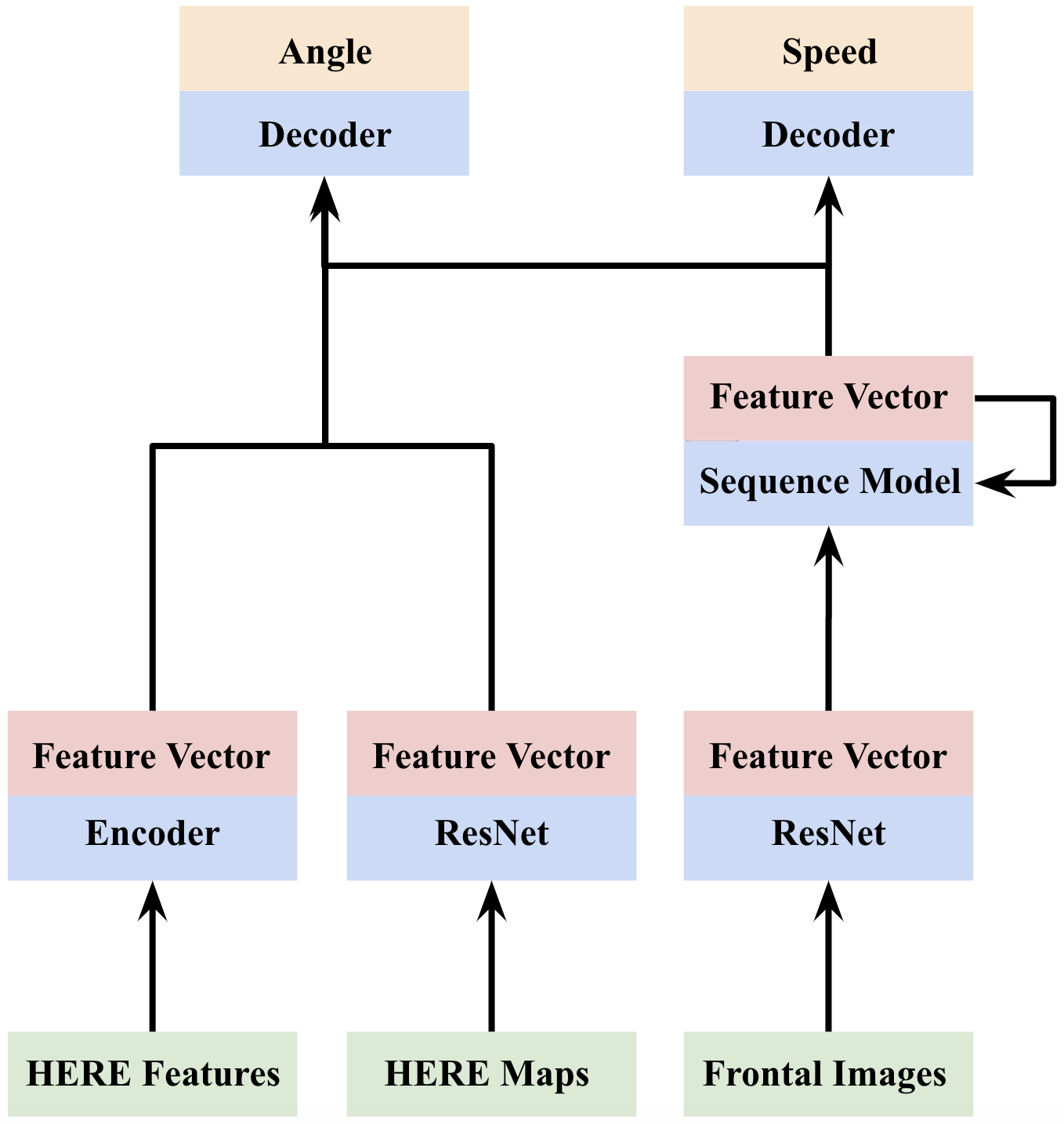}
    \caption{Model 3: The network consists of a pre-trained ResNet and fully connected network that feeds into an LSTM model for frontal images input, a pre-trained ResNet and fully connected network for HERE map images input. Each regressor consists of 3 blocks of a linear layer, a ReLU activation, and a 20\% dropout. The hidden layers are 64, 32, and 1.}
    \label{fig:model3}
\end{figure}

Figure \ref{fig:model3} shows Model 3's network architecture. The inputs consist of the front facing GoPro \cite{gopro} images, the semantic map images and numerical features from HERE Technologies \cite{here}. The model uses a pre-trained ResNet for both frontal images and HERE maps, while the semantic map features are passed through a fully connected network. The output feature vector of the frontal images is fed into an LSTM model. Finally, the output of the LSTM and the output of the ResNet models are concatenated along with HERE numerical features (semantics) and fed into the angle and speed regressor.

\subsection{Implementation}

\subsubsection{Model 1 implementation details (1st place)}

\paragraph{Hyper-parameters for network training.}
All models are trained using the Adam optimizer with an initial learning rate of 0.0001, without weight decay, and momentum with $\beta_1 = 0.9$, and $\beta_2 = 0.999$. In all model variants we optimize the sum of mean squared error (MSE) for the steering angle and speed. Minibatch sizes are 8, 32, or 64, limited by GPU memory. The image set, depending on run, is of dimensions 320x180 or 160x90 using ResNet34 or ResNet152 models. Table \ref{tab:results_model1} summarizes the hyperparameters and image settings used for each model 1 variant.

\paragraph{Ensemble.} \label{ensemble}
We quantize the steering angles and speeds from the training dataset into 100 and 30 evenly spaced bins respectively and use these discrete distributions to compute a weighted average of predicted values. Table \ref{tab:results_model1} shows the performance of each individual model 1 variant and the ensemble.

\paragraph{Computation.} 
Preprocessing time is around 3 days of computation on an Intel i7-4790k CPU using 6 cores with data stored on a 7200rpm hard drive. The largest bottleneck in preprocessing is the I/O due to the large number of image. Models 1 variants were trained using a single Nvidia GPU. Each epoch required approximately 10-12 hours.


\subsubsection{Model 2 implementation details (2nd place)}

\paragraph{Model 2-single.} This model uses the Adam optimizer with an initial learning rate of 0.0003. We implemented learning rate decay to 0.0001, 0.00005, and 0.00003 after the first 5, 15, and 20 epochs. Overall the model was trained for 90 epochs, with a batch size of 13 for training, validation and testing. The loss criterion used for both speed and steering angle was MSE loss, and the overall model loss was defined as the summation of the speed loss and steering angle loss. Model 2-stacked was trained in an identical manner, but the lowest MSE loss was achieved after the 14th epoch. We used a similar method to train all model 2 variants, with slight changes in learning rate and learning rate decay tuned over multiple runs. Our models did not require many epochs to reach their lowest validation MSE, and an ensemble of the results of the individual models outperformed each individually. 

\paragraph{Model 2-sequence.} The model is trained using the Adam optimizer with an initial learning rate of 0.003, which is halved after epochs 20, 30 and 40. We use the same combination loss as in model 2-single and model 2-stacked, summing the MSE of speed and steering angle. Taking an ensemble of model 2 variants decreased the MSE for speed from 6.115 to 5.312, and the total MSE for steering from 925 to 901, achieving 2nd place overall in the competition. 

\paragraph{Hyper-parameters for network training.}
In each training run of the model, we save only the best model for speed and the best model for steering angle, which were then separately stored and used for predicting values on the test set. 

\subsubsection{Model 3 implementation details (3rd place)}
We used the HERE map images, HERE numerical features, and frontal images as inputs. In addition to the baseline mode, we passed the HERE map images into a pre-trained ResNet, and concatenated its output with the output from encoded frontal images and HERE numerical features. The concatenated feature vector is decoded by fully connected networks for predicting steering angle and speed. 

\paragraph{Hyper-parameters for network training.}
The model is trained using an Adam optimizer with learning rate 0.0001. We use a ResNet34 for the HERE MAP images, and ResNet50 for frontal images. We apply dropout with 0.2 and 0.5 probabilities and batch size of 64 for training. 

\paragraph{Computation.} 
Pre-processing time is around 6 hours on a AMD 2700X CPU with data stored on a 7200rpm hard drive. The model is trained using an Nvidia GTX 1080. Each epoch requires approximately 5 minutes to complete.
\section{Results}
\label{sec:results}

\begin{figure*}
    \begin{center}
        \includegraphics[width=0.195\linewidth]{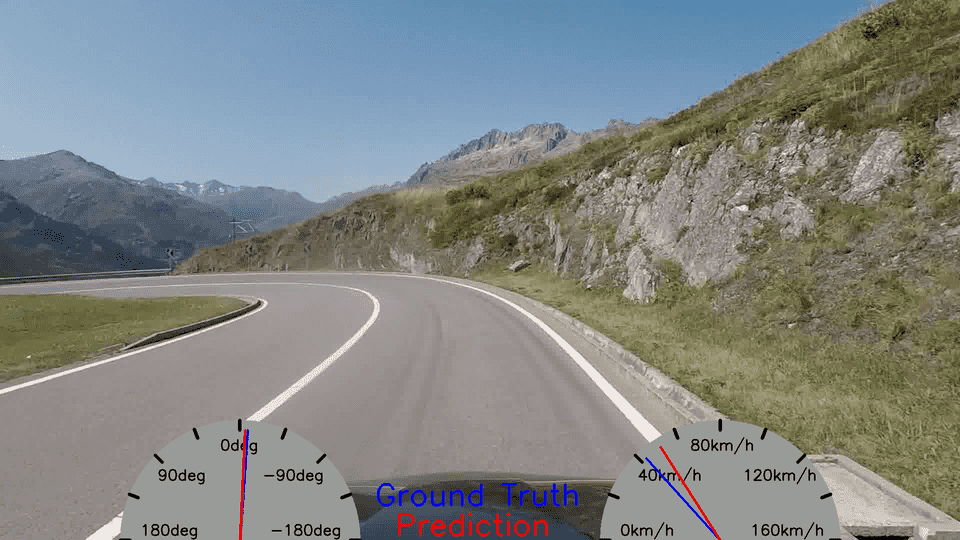}
        \includegraphics[width=0.195\linewidth]{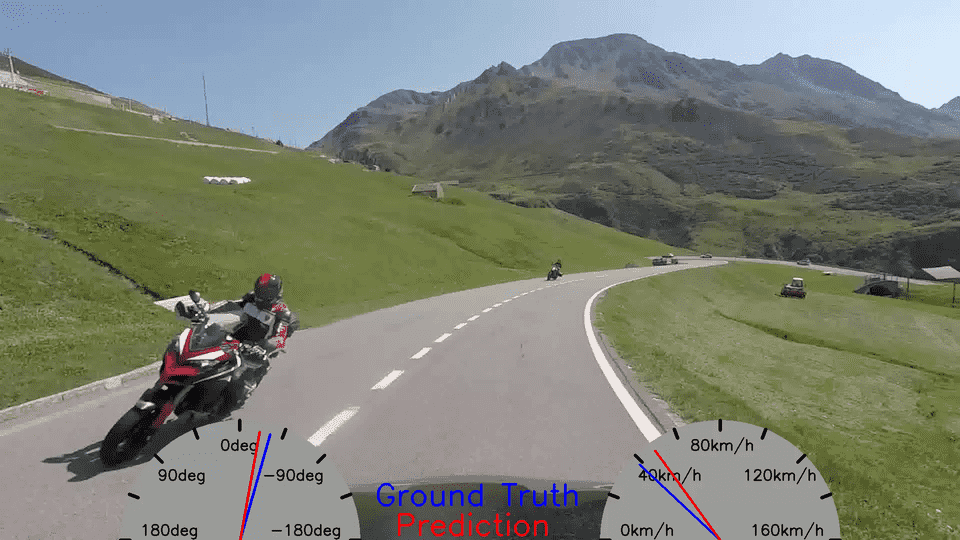}
        \includegraphics[width=0.195\linewidth]{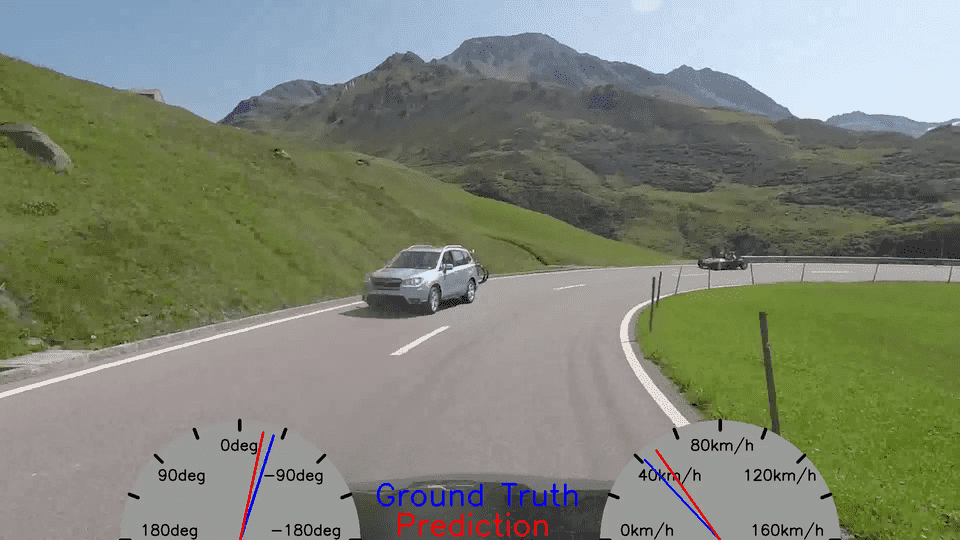}
        \includegraphics[width=0.195\linewidth]{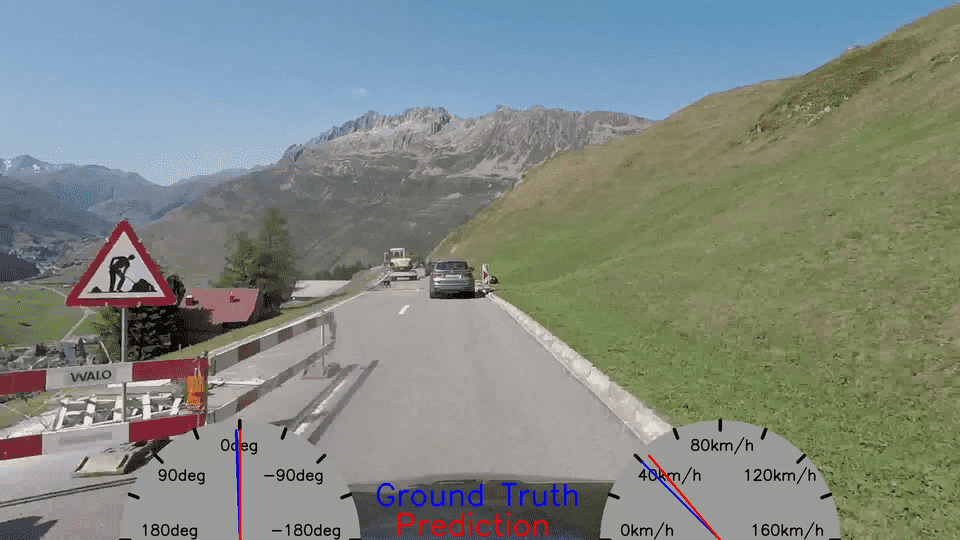}
        \includegraphics[width=0.195\linewidth]{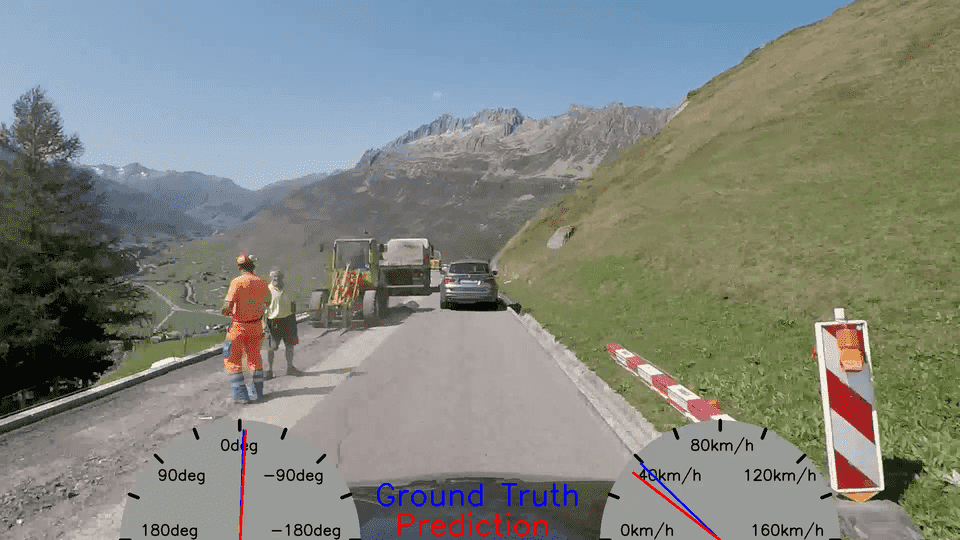}
        \includegraphics[width=0.195\linewidth]{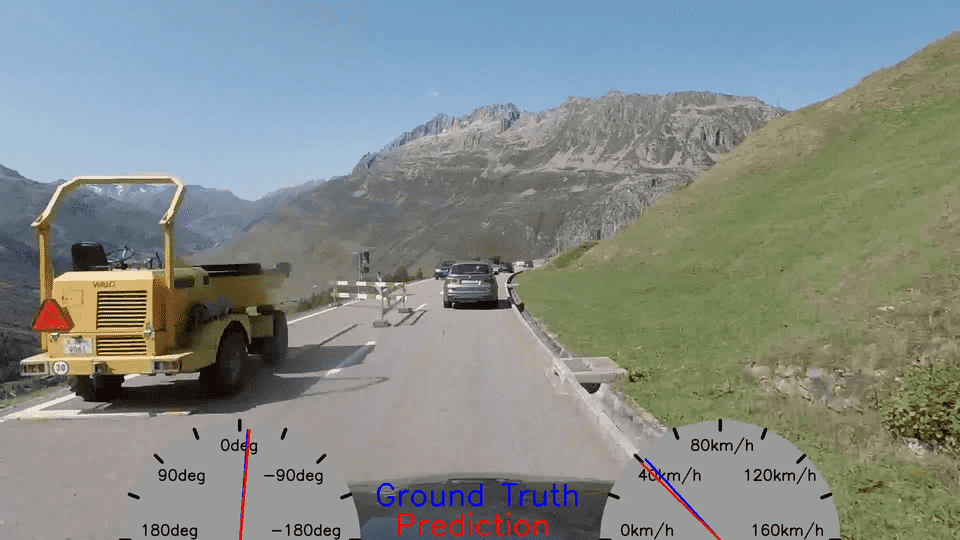}
        \includegraphics[width=0.195\linewidth]{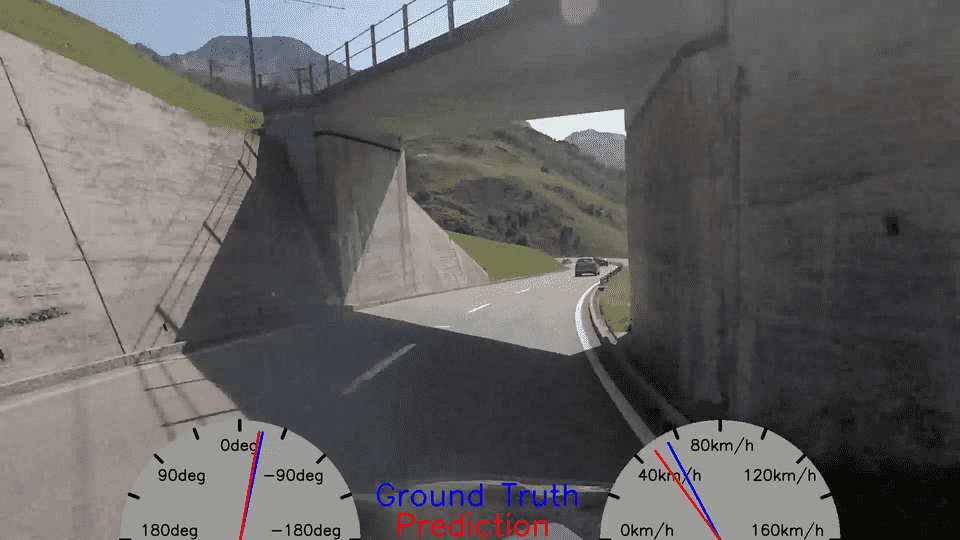}
        \includegraphics[width=0.195\linewidth]{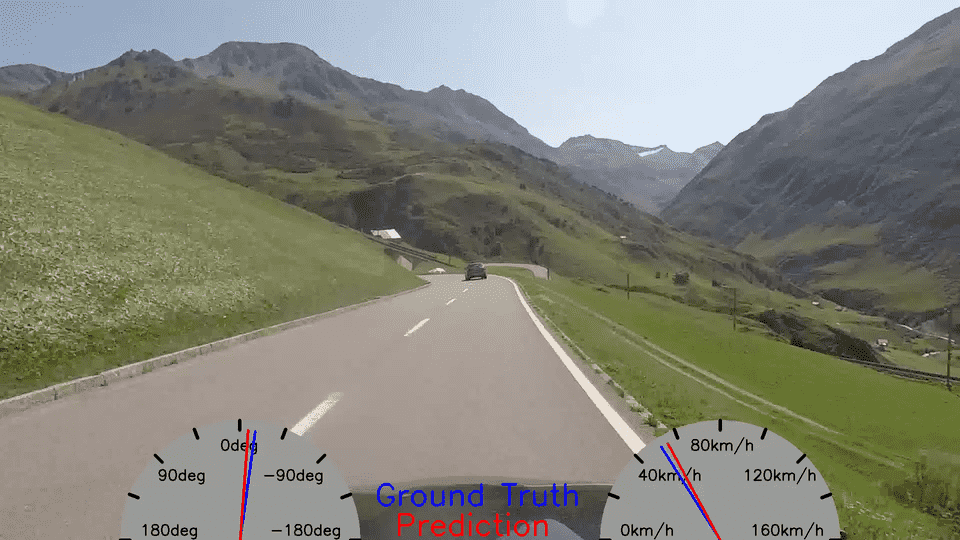}
        \includegraphics[width=0.195\linewidth]{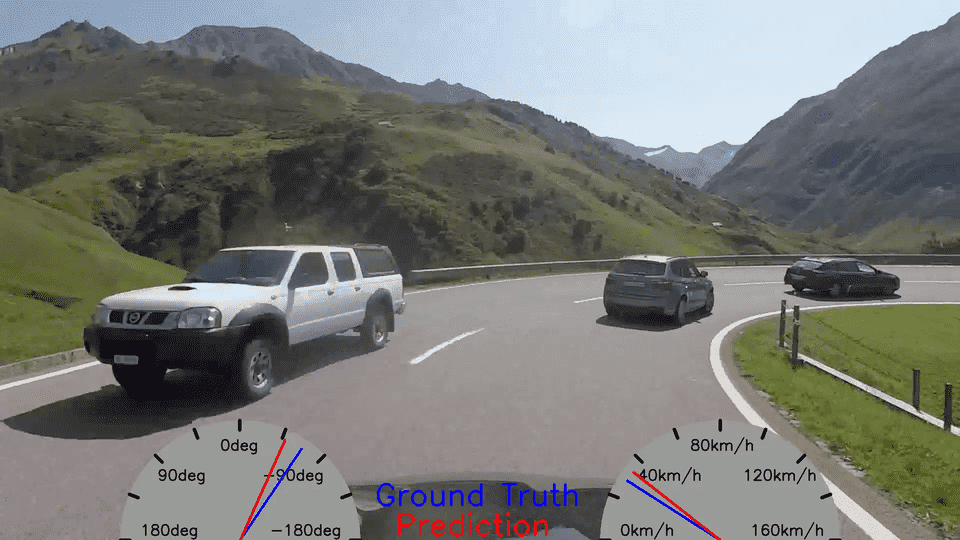}
        \includegraphics[width=0.195\linewidth]{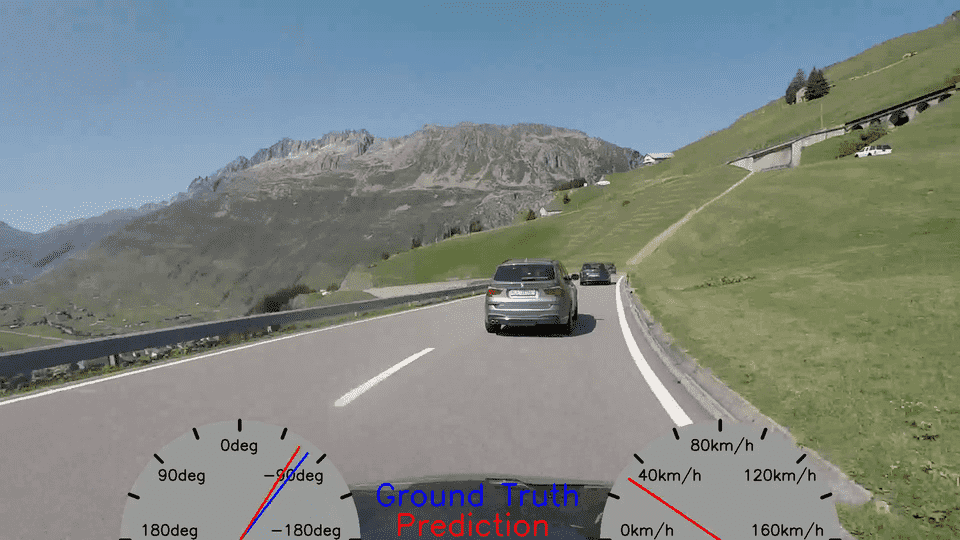}
        \end{center}
    \caption{Sample of test results with the ground truth (in blue) and predicted values (in red) for steering angle and driving speed. The ground truth and the prediction range between $-180\degree$ to $180\degree$ for steering angle and between $0$ to $160\text{km/h}$ for driving speed.}
    \label{fig:predictions}
\end{figure*}

\begin{figure*}
\begin{center}
    \includegraphics[width=0.495\linewidth]{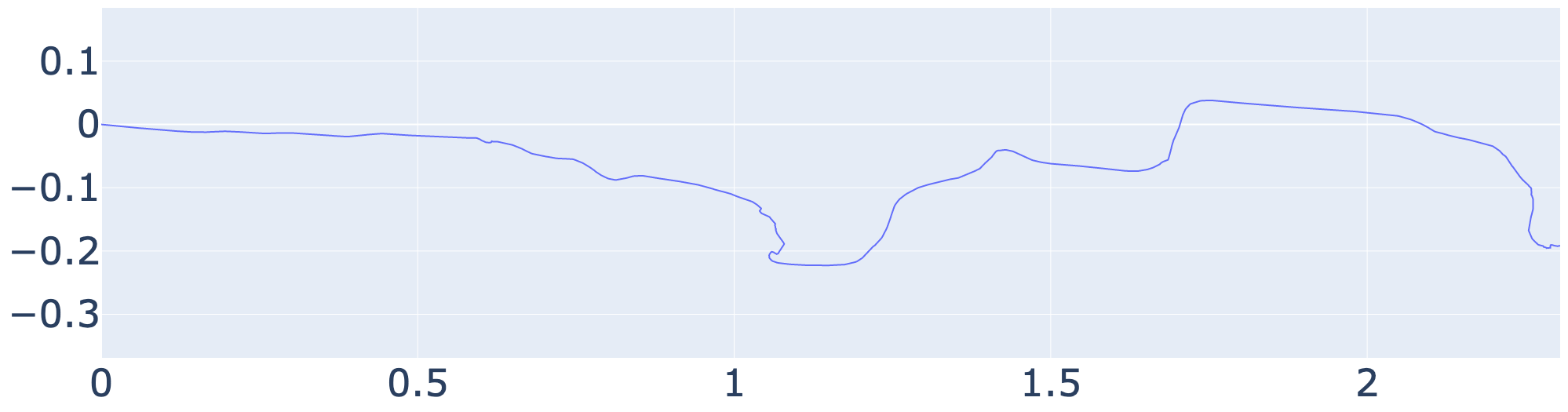}
    \includegraphics[width=0.495\linewidth]{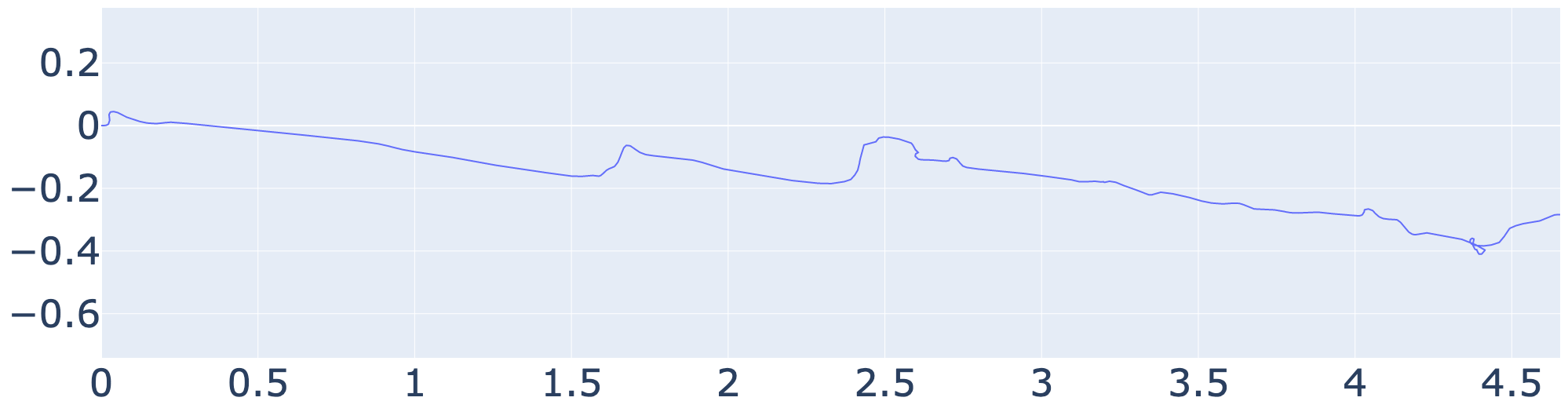}
    \caption{Sample predicted test paths: The vertical and horizontal axes represent distance in kilometers. Each trajectory starts at the zero coordinates and moves in the predicted direction and speed at each time point. The predictions for both steering angle and driving speed are 100 milliseconds apart and the trajectory is calculated with constant angle and speed between measurements.}
    \label{fig:path}
\end{center}
\end{figure*}

\begin{figure}
\begin{center}
    \includegraphics[width=\linewidth]{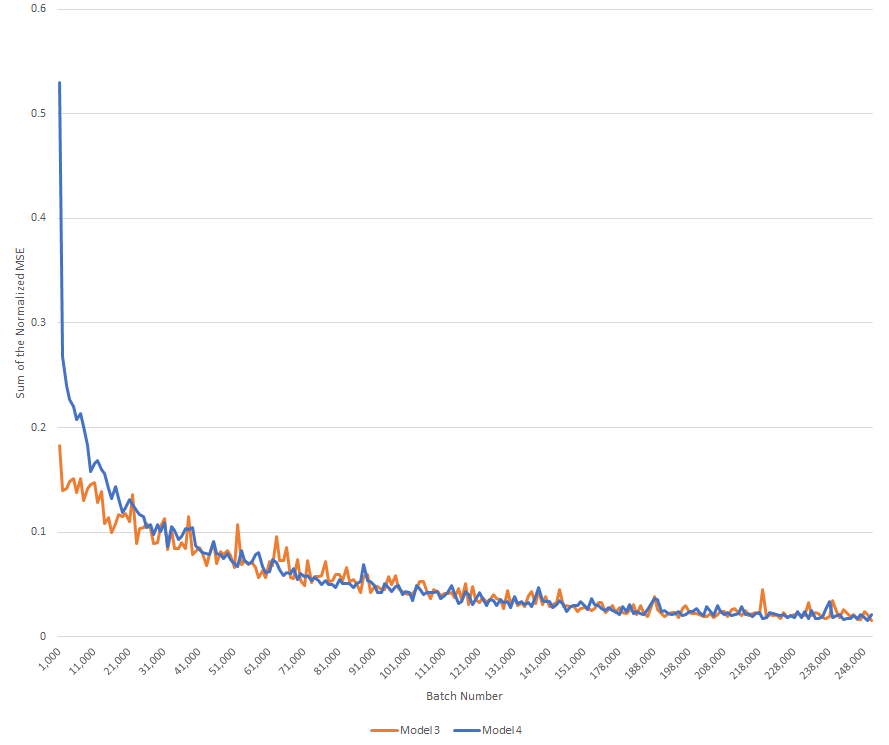}
    \caption{Training loss of angle and speed MSE as a function of epoch.}
    \label{fig:trainloss}    
\end{center}
\end{figure}

We have won the ICCV 2019 Learning to Drive Challenge \cite{learning2drive} in all first three places overall, compared with other previous state-of-the art positioned below 10th place. Figure \ref{fig:predictions} shows a sample of test prediction results.
Ground truth angle and speed are in blue and predicted angle and speed are in red. The entire video is available online \cite{winningvideo}. 
Figure \ref{fig:path} show the predicted path for two driving chapters, where the axis scales represent distance in kilometers. 
Table \ref{tab:results_model1} summarizes the results of model 1 which won 1st place overall by fusing together different modalities using a neural network.
We train each of our five neural network model 1 variants for varying number of epochs to make efficient use of our computational resources. 
Table \ref{tab:results_model1} shows that the most significant improvement on the single models occurs by including of the HERE semantic map into the model, which results in a decrease of the total MSE by approximately 300 points on the Angle metric. 
This is most apparent between model 1 variants 1 and 4. Including the city location as part of the semantic map has a marginal benefit to the model, as seen by the change from model 1 variants 4 to 5. 
Notably, models 1 variants 3 and 4 tend to overfit. Our best result for both is after 2 epochs, and the MSE slowly increases for each additional epoch. The training loss for both models decrease throughout the training, as shown in Figure \ref{fig:trainloss}. The performance of the ensemble method in the different road types is shown in Table \ref{tab:mse}.

\begin{figure}
\begin{center}
    \includegraphics[width=1\linewidth]{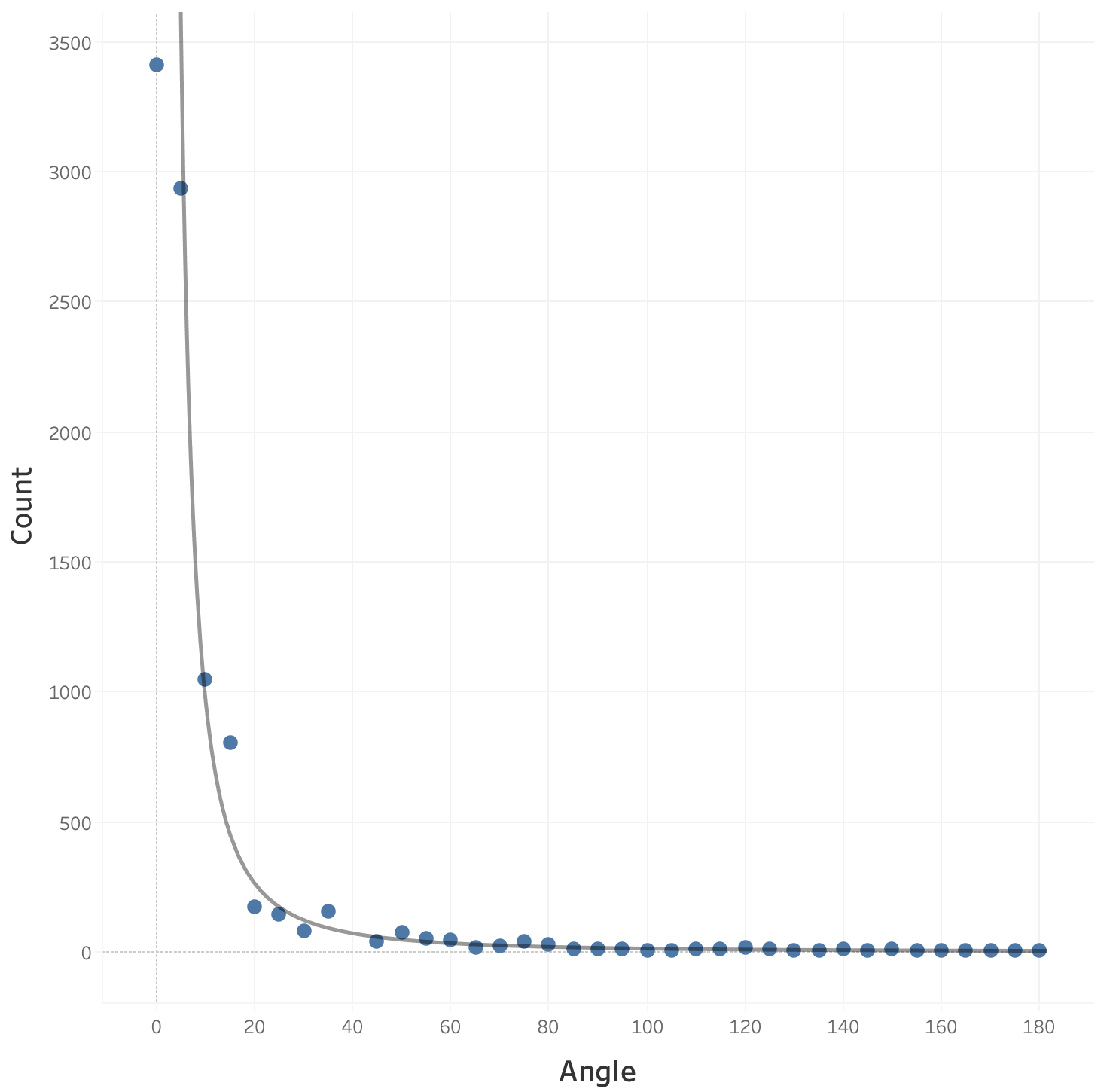}
    \caption{The distribution of angles between 0 and 180 in absolute values for bins of 5 angles each.}
    \label{fig:angle_count}
\end{center}
\end{figure}

\begin{figure}
\begin{center}
    \includegraphics[width=1\linewidth]{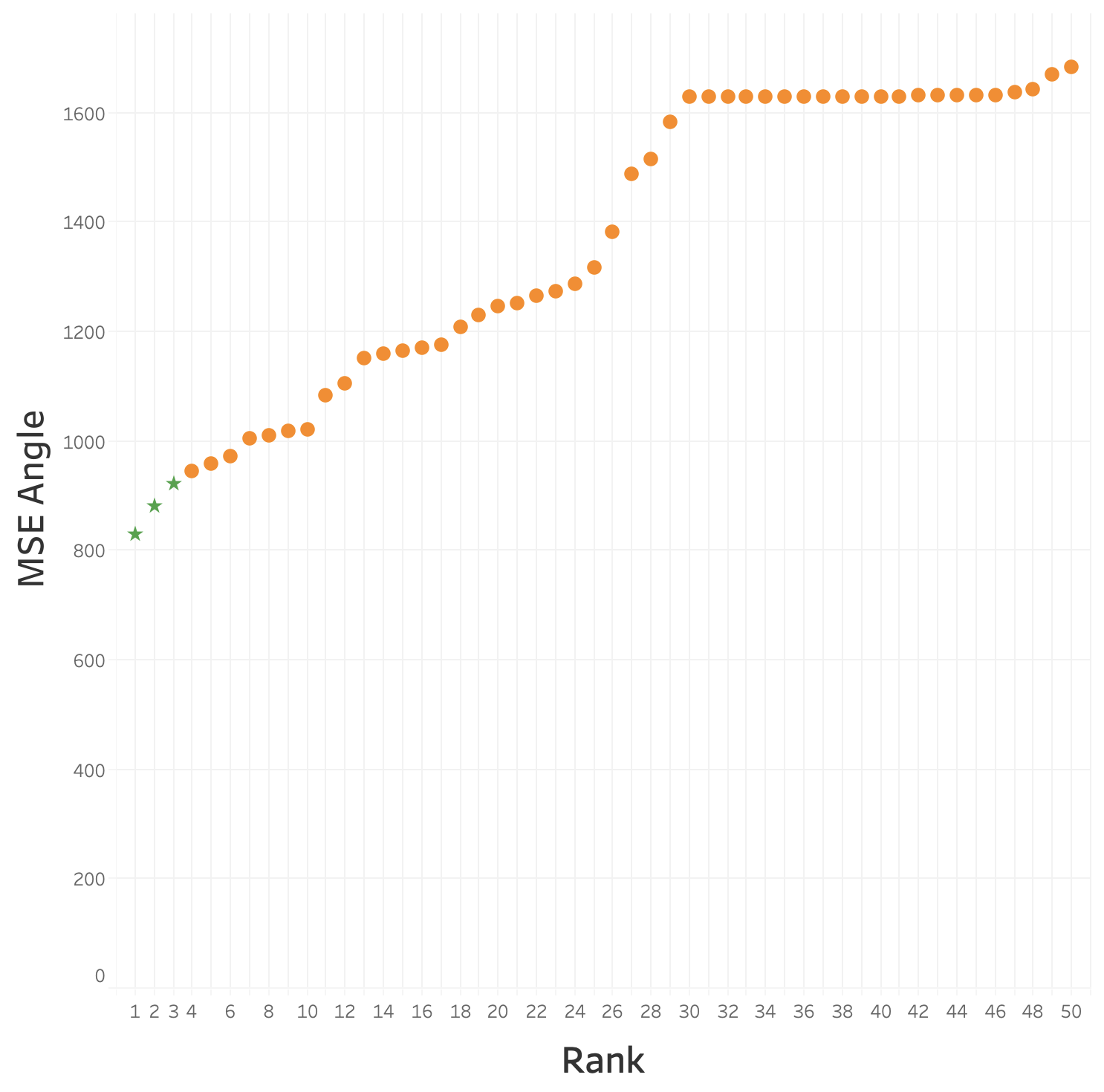}
    \caption{Top 50 teams ranked by total MSE angle. Our methods are the top 3 performers as denoted by stars. The total MSE angle achieved by the models are Model 1: 831.424, Model 2: 881.513 and Model 3: 922.836.}
    \label{fig:top50}
\end{center}
\end{figure}

Table \ref{tab:results_model2} summarizes the results of model 2 which won 2nd place overall by augmenting the data with it segmentation maps.
Figure \ref{fig:angle_count} shows the distribution of angles between 0 and 180 in absolute values for bins of 5 angles each. As expected, the number instances in each angle decreases exponentially as angle is increased. Finally, Figure \ref{fig:top50} show the total MSE for the top 50 rankes teams, with our models achieving the best performance overall, winning 1st, 2nd, and 3rd.

\begin{table*}
\small
\begin{center}
\begin{tabular}{lccccc|ccc}
Model    & CNN       & Dimensions & Semantic Map & Batch & Epochs & MSE Angle & MSE Speed & Combined \\
\hline
1        & ResNet34  & 320x180          & No           & 64    & 2      & 1111.437 & 5.866 & 1117.303 \\
\hline
2        & ResNet152 & 320x180          & No           & 8     & 1      & 1211.434 & 5.461 & 1216.895 \\
\hline
3        & ResNet34  & 160x90           & 20 Features  & 8     & 1      & 897.489  & 6.664 & 904.153  \\
         &           &                  &              &       & \textbf{2}      & \textbf{883.501}  & \textbf{6.403} & \textbf{889.904}  \\
         &           &                  &              &       & 3      & 931.689  & 6.445 & 938.134  \\
         &           &                  &              &       & 4      & 970.96   & 6.714 & 977.674  \\
         &           &                  &              &       & 5      & 956.262  & 6.576 & 962.838  \\
\hline
4        & ResNet34  & 320x180          & 20 Features  & 32    & 1      & 995.42   & 5.316 & 1000.736 \\
         &           &                  &              &       & 2      & 946.516  & 5.337 & 951.853  \\
         &           &                  &              &       & 3      & 989.013  & 5.519 & 994.532  \\
         &           &                  &              &       & 4      & 965.791  & 5.706 & 971.497  \\
         &           &                  &              &       & 5      & 987.572  & 5.846 & 993.418  \\
\hline
5        & ResNet34  & 320x180          & 47 Features  & 64    & 1      & 900.407  & 5.571 & 905.978  \\
\hline
Ensemble &           &                  &              &       &        & \textbf{831.504}  & \textbf{4.543} & \textbf{836.047} \\
\hline
\end{tabular}
\end{center}
\caption{Results for model 1 variants and their ensemble: The best overall result is an ensemble of the single models. Individually, we note that the inclusion of the semantic map reduces the MSE by about 300 (comparing models 1 and 4) and using the smaller image size resulted in an additional benefit (comparing models 3 and 4). Models 3 and 4 likely suffer from overfitting as evidence by the increasing test MSE, although the training loss decreased, as shown in Figure \ref{fig:trainloss}. The best standalone and overall models are in bold.}
\label{tab:results_model1}
\end{table*}

\begin{table}
\small
\begin{center}
\begin{tabular}{l|cc}
Model 1         & MSE Angle & MSE Speed \\
\hline
Overall       & 831.5    & 4.5   \\
Zone30        & 2,981.1  & 0.3   \\
Zone50        & 1,353.4  & 6.0   \\
Zone80        & 168.6    & 4.1   \\
Right         & 1,928.4  & 1.3   \\
Straight      & 821.7    & 4.6   \\
Left          & 833.6    & 2.6   \\
Pedestrian    & 3,722.1  & 4.1   \\
Traffic Light & 329.2    & 5.3   \\
Yield         & 1,818.9  & 2.8  \\
\hline
\end{tabular}
\end{center}
\caption{Performance across various zones: the ensemble method did worst in the pedestrian sections and best in Zone80. Presumably, pedestrian sections would be hardest to train due to the unpredictability of cities and people. Zone80 sections are likely straighter and require less change in speed and steering angle and would probably be easier to train. Likewise, Right and Left sections, which would require learning a turn, would be difficult, but Straight segments are easier to train and did better. }
\label{tab:mse}
\end{table}

Model 2-sequence achieved the lowest MSE on the test set at 6.115 while the best performing single model for steering angle was model 2-stacked with total MSE loss at 925.926. The results are summarized in Table \ref{tab:results_model2}. Model 2-stacked performed the best for steering as a result of the combination of data augmentation which included horizontal flipping of images in a sequence, and the full view of the sequence to the input of the model.

\begin{table}
\small
\begin{center}
\begin{tabular}{l|cc}
Model 2 & MSE Speed & total MSE Angle \\ 
\hline
Model 2-single &7.440&\multicolumn{1}{r}{1,140.875}\\ 
Model 2-stacked &7.036&\multicolumn{1}{r}{925.926}\\ 
Model 2-sequence  &6.115&\multicolumn{1}{r}{1,075.497}\\ 
$\textbf{Avg:}$ & $\textbf{5.312}$ & \multicolumn{1}{r}{$\textbf{901.802}$}  \\ 
\hline
\end{tabular}
\end{center}
\caption{Results of model 2: The stacked variant achieved the lowest total MSE for steering and was second for speed. The individual model that had the best performance for speed was model 2-sequence. The average of the predictions generated by model 2-stacked and model 2-sequence provided a significant decrease in the total MSE of both speed and steering angle. This combination is the final result of model 2 and it was our best submission to the competition which awarded us the second place in steering angle prediction and second place overall.}
\label{tab:results_model2}
\end{table}

\section{Conclusions}
\label{sec:conclusions}
In conclusion, our main contributions are (i) fusing together multiple modalities, (ii) augmenting the inputs with segmentation maps generated from a pre-trained model, (iii) leveraging the fact that the output is a smooth function, and an ensemble of diverse model architectures to win the ICCV Learning to Drive Challenge. We demonstrate high quality steering angle and speed predictions to yield accurate driving trajectories. In the spirit of reproducible research we make our models and code publicly available.

\newpage
\clearpage

{\small
\bibliographystyle{ieee_fullname}
\bibliography{egbib}
}

\end{document}